\documentclass{article}

\usepackage[preprint]{neurips_2025}

\usepackage[utf8]{inputenc} 
\usepackage[T1]{fontenc}    
\usepackage{hyperref}       
\usepackage{url}            
\usepackage{booktabs}       
\usepackage{amsfonts}       
\usepackage{nicefrac}       
\usepackage{microtype}      
\usepackage{xcolor}         
\usepackage{enumitem} 		
\usepackage[capitalize]{cleveref} 
\usepackage{graphicx} 		
\usepackage{caption}
\usepackage{subcaption}
\usepackage{float}
\usepackage{wrapfig}
\usepackage{multicol}

\title{SNAP: A Benchmark for Testing the Effects of Capture Conditions on Fundamental Vision Tasks}

\author{%
  Iuliia Kotseruba\\
  York University \\
  \texttt{yulia84@yorku.ca} \\
  \And
  John K. Tsotsos \\
  York University \\
  \texttt{tsotsos@yorku.ca} \\
}

\begin{document}

\maketitle

\begin{abstract}
Generalization of deep-learning-based (DL) computer vision algorithms to various image perturbations is hard to establish and remains an active area of research. The majority of past analyses focused on the images already captured, whereas effects of the image formation pipeline and environment are less studied. In this paper, we address this issue by analyzing the impact of capture conditions, such as camera parameters and lighting, on DL model performance on 3 vision tasks---image classification, object detection, and visual question answering (VQA). To this end, we assess capture bias in common vision datasets and create a new benchmark, SNAP (for \textbf{S}hutter speed, ISO se\textbf{N}sitivity, and \textbf{AP}erture), consisting of images of objects taken under controlled lighting conditions and with densely sampled camera settings. We then evaluate a large number of DL vision models and show the effects of capture conditions on each selected vision task. Lastly, we conduct an experiment to establish a human baseline for the VQA task. Our results show that computer vision datasets are significantly biased, the models trained on this data do not reach human accuracy even on the well-exposed images, and are susceptible to both major exposure changes and minute variations of camera settings.

\end{abstract}

\section{Introduction}

Data is one of the pillars of modern deep-learning-based (DL) computer vision as both training and evaluation of algorithms rely heavily on large image datasets. Generalization beyond training data is highly desirable but difficult to determine for current DL models \cite{nagarajan2019uniform, dziugaite2020search, zhang2021understanding, chatterjee2022generalization}. Several factors contribute to this, including DL model opacity combined with lack of established DL theory \cite{goldblum2020truth, he2020recent, suh2024survey} and the growing volume of training data \cite{qin2024dataset}. Past research on DL generalization focused on measuring and mitigating data biases and sensitivity of DL models to input perturbations, such as adversarial examples \cite{wei2024physical}, image distortions \cite{hendrycks2019benchmarking}, viewing angles \cite{barbu2019objectnet}, incongruous context \cite{hendrycks2021natural}, etc. So far, most works focused on images already captured but not the image formation process, which has many variables that depend on the sensor, its settings, and environment properties. If any of these elements are altered, images of the same scene may appear very differently. However, the effects of these changes on vision algorithms have not been systematically examined.

In this paper, we focus on the effects of capture conditions, i.e. camera parameters and lighting, on 3 fundamental visual tasks---image classification, object detection, and visual question answering (VQA). To do so, we 1) analyze capture bias in the existing vision datasets; 2) gather a new dataset, SNAP, where capture bias is minimized by dense sampling of camera parameters under controlled lighting conditions; 3) estimate the effects of capture settings on vision tasks by testing a representative set of models on SNAP; 4) associate model performance with the properties of the training data and models themselves; and 5) conduct a human study to establish a human baseline for common vision tasks under different capture conditions. 

\section{Related Works}

\noindent
\textbf{Dataset biases.} Most vision datasets are biased, which affects the representations and generalization ability of the trained algorithms. Classification datasets, in particular, are found to be biased in terms of object classes, sizes, backgrounds, and viewpoints \cite{ponce2006dataset,torralba2011unbiased, herranz2016scene, tommasi2017deeper, azylay2018why, tsotsos2021probing, zeng2024understanding, liu2025decade}. To our knowledge, only a few works considered sensor parameter bias (referred to as \textit{capture bias} in \cite{tommasi2017deeper}) in computer vision data. These include studies of consumer photos \cite{wueller2008statistic,wueller2018statistic}, analyses of VOC2007 \cite{everingham2010pascal} and COCO \cite{lin2014microsoft} metadata in \cite{tsotsos2019does}, and common Exif tags for YFCC100M \cite{thomee2016yfcc100m} reported in \cite{zheng2023exif}. 

\noindent
\textbf{Effect of image degradations on algorithms.} A large number of studies investigated the effect of artificial image degradations on the performance of classical and DL vision and vision-language algorithms. To facilitate this, widely used datasets have been modified with artificial corruptions and perturbations, starting with ImageNet-C \cite{hendrycks2019benchmarking}, followed by MNIST-C \cite{mu2019mnist}, as well as Pascal-C, COCO-C, and Cityscapes-C \cite{michaelis2019benchmarking}.  Some of the common distortions considered in prior works are blur \cite{dodge2016understanding,dodge2017study, grm2017strengths,roy2018effects,che2019gazegan,hendrycks2019benchmarking,chen2023benchmarking,yamada2022does}, various types of noise \cite{le2011robustness,dodge2016understanding,liu2016evaluation,niu2016evaluation, rodner2016fine,tow2016robustness,rodner2016fine,dodge2017study,grm2017strengths,roy2018effects, geirhos2018generalisation,hendrycks2019benchmarking,chen2023benchmarking,yamada2022does}, digital artifacts \cite{le2011robustness,dodge2016understanding,zheng2016improving,roy2018effects, che2019gazegan,hendrycks2019benchmarking,chen2023benchmarking,yamada2022does}, changes in brightness \cite{tow2016robustness, grm2017strengths, hendrycks2019benchmarking}, contrast \cite{dodge2016understanding, grm2017strengths, geirhos2018generalisation, che2019gazegan, hendrycks2019benchmarking}, colour \cite{rodner2016fine, geirhos2018generalisation}, etc.

While it was noted that blur, noise, and other distortions may also be caused by changing camera settings \cite{tow2016robustness, roy2018effects, bielova2019digital, che2019gazegan}, only a few studies considered the effects of natural capture conditions. One of the early tests \cite{andreopoulos2011sensor} demonstrated performance fluctuations of classical vision algorithms on 4 scenes acquired with different shutter and gain settings under 3 illumination conditions. In \cite{wu2017active}, classical and DL object detectors were tested on the set of 5 objects taken with 64 camera configurations and 7 lighting levels. Another study evaluated DL object detectors on subsets of COCO corresponding to different camera settings and showed performance variations on bins with different number of training samples \cite{tsotsos2019does}. Most recently, the authors of \cite{baek2024unexplored} collected a new dataset, ImageNet-ES by displaying 2000 images from the original ImageNet on the TV screen and photographing each across 64 camera parameter sets with and without external lighting. The results showed the sensitivity of all tested models to these changes and benefits of diversifying training data w.r.t. capture conditions.

\noindent
\textbf{Effect of image degradations on human vision.} Several studies compared human and machine performance on various types of image degradations and tasks (e.g., fine-grained image classification \cite{dodge2017study}, saliency \cite{che2019gazegan}, and object recognition \cite{geirhos2018generalisation, shen2025assessing}). Performance of both humans and CNNs was shown to be affected by various image degradations but to a different extent. Human performance declined significantly only for the highest distortion levels \cite{geirhos2018generalisation, shen2025assessing}. And while some deep networks outperformed humans on undistorted images (e.g., on fine-grained dog breed classification \cite{dodge2017study}) they degraded faster under distortions \cite{geirhos2018generalisation, shen2025assessing}. Both humans and CNNs were reported to be relatively robust to minor color-related distortions (greyscale conversion or opponent colors) \cite{geirhos2018generalisation}. 

This paper expands previous work both in depth and scope: 1) we analyze Exif data of 13 common vision datasets, particularly focusing on those used for training and evaluation of models for object detection and classification and foundation vision models, 2) collect a dataset of 100 real scenes with over 700 densely sampled camera parameter combinations under 2 controlled illumination conditions, 3) test 52 models on classification, detection, and VQA tasks and collect data from 43 subjects for the latter task, and 4) systematically analyze performance of algorithms w.r.t. capture conditions, compare it to humans, and link it to biases in the training data.

\section{Methodology}

This session discusses the paradigm, procedure, metrics, and models for probing DL vision algorithms across capture conditions. In \cref{sec:dataset_analysis}, we show significant capture bias in common vision datasets that motivates the balanced design of the SNAP dataset described in \cref{sec:SNAP_dataset}. \cref{sec:human_exp} is dedicated to the human experiment conducted on a subset of SNAP to establish a human baseline for vision tasks under varying capture conditions. Lastly, \cref{sec:models_and_metrics} lists the models and metrics chosen for evaluation. Additional results and details are available in the Appendix.

\subsection{Analysis of capture bias in common vision datasets}
\label{sec:dataset_analysis}

To analyze capture bias in vision data, we chose popular datasets for training image classifiers, object detectors, and foundation models. These include VOC2007 \cite{everingham2010pascal}, ImageNet \cite{deng2009imagenet}, SBU \cite{ordonez2011im2text}, COCO \cite{lin2014microsoft}, YFCC15M \cite{thomee2016yfcc100m}, CC3M \cite{sharma2018conceptual}, ImageNet21K \cite{ridnik2021imagenet}, CC12M \cite{changpinyo2021conceptual}, WIT \cite{Srinivasan2021WIT}, LAION400M \cite{schuhmann2021laion}, OpenImages v7 \cite{benenson2022colouring}, COYO \cite{kakaobrain2022coyo-700m}, and Wukong \cite{gu2022wukong}, with over 1B images combined. 

\begin{wraptable}[16]{R}{0.55\textwidth}
\centering
\vspace{-1em}
\caption{Availability of images and metadata in the common vision datasets. ``w/ Exif''---all 3 tags of interest (shutter speed, F-number, and ISO) are in metadata.}
\label{tab:exif_stats}
\resizebox{0.55\textwidth}{!}{%
\begin{tabular}{@{}lcccccc@{}}
\toprule
\textbf{Dataset}       & \textbf{Year} & \textbf{\# images} & \textbf{\# downloaded} & \textbf{\# w/ Exif} \\ \midrule
VOC2007 \cite{everingham2010pascal} & 2007 & 10.0K     & 10.0K (100\%)   & 3.0K (29.7\%) \\
ImageNet \cite{deng2009imagenet} & 2009 & 1.6M      & 1.6M (100\%)   & 59.7K (3.8\%) \\
SBU \cite{ordonez2011im2text} & 2011 & 1.0M      & 749.3K (74.9\%)    & 492.3K (49.2\%) \\
COCO \cite{lin2014microsoft} & 2014 & 164.0K    & 164.0K (100\%)   & 68.6K (41.8\%) \\
OpenImages v7 \cite{benenson2022colouring} & 2016 & 1.9M      & 1.9M (100\%)   & 1.1K (0.1\%) \\
YFCC15M \cite{thomee2016yfcc100m} & 2016 & 15.4M     & 13.0M (84.6\%)    & 9.5M (61.6\%) \\
CC3M \cite{sharma2018conceptual} & 2018 & 3.3M      & 2.3M (70.0\%)    & 105.1K (3.2\%) \\
ImageNet21K \cite{ridnik2021imagenet} & 2021 & 13.2M     & 13.2M (100\%)   & 797.0K (6.1\%) \\
CC12M \cite{changpinyo2021conceptual} & 2021 & 12.4M     & 8.1M (64.9\%)    & 280.5K (2.3\%) \\
WIT \cite{Srinivasan2021WIT} & 2021 & 26.5M     & 25.9M (97.8\%)    & 12.1M (45.7\%) \\
LAION400M \cite{schuhmann2021laion} & 2021 & 413.6M    & 304.4M (73.6\%)    & 5.8M (1.4\%) \\
Wukong \cite{gu2022wukong} & 2022 & 101.4M    & 96.0M (95.0\%)    & 125.2K (0.1\%) \\
COYO \cite{kakaobrain2022coyo-700m} & 2022 & 747.0M    & 512.9M (68.7\%)    & 15.3M (2.1\%) \\ \midrule
Total         & -    & 1.3B      & 980.1M (73.3\%)    & 44.7M (3.3\%) \\ \bottomrule
\end{tabular}%
\vspace{-2em}
}
\end{wraptable}

\cref{tab:exif_stats} shows a summary of the datasets and metadata availability. Overall, metadata in the datasets is distributed highly unevenly; datasets originating from the Common Crawl generally contain much fewer images with metadata (1-6\%) than datasets from curated resources, such as Flickr and Wikipedia (30-60\%). 

\begin{wrapfigure}[16]{r}{0.55\textwidth}
\centering
\vspace{-3em}
\includegraphics[width=0.55\textwidth]{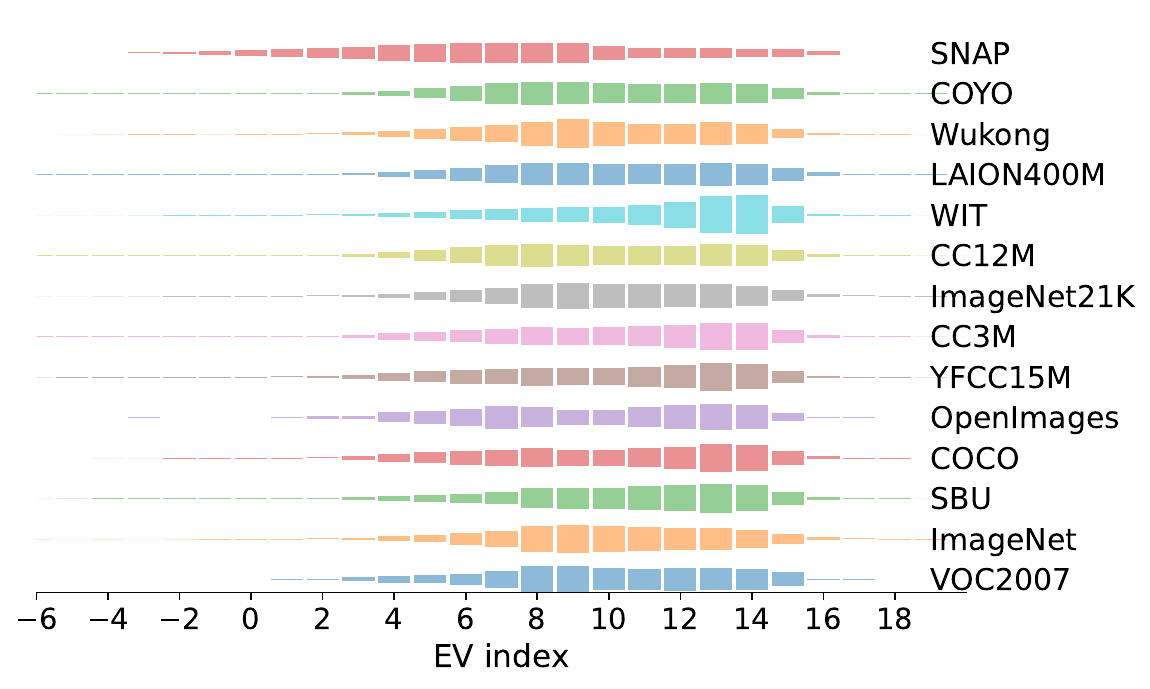}
\caption{Normalized distribution of camera settings grouped by EV index for each dataset. Plots are arranged chronologically from bottom to top. Our SNAP dataset introduced in \cref{sec:SNAP_dataset} is shown for reference.}
\label{fig:datasets_EV_idx_dist}
\end{wrapfigure}

We extracted Exif information from images belonging to these datasets. We used the Flickr API to gather metadata for datasets with images from Flickr. For the rest, we downloaded the images and extracted Exif tags using the ExifTool \cite{exiftool}. Because many recent datasets distribute only image URLs, not all images could be downloaded and analyzed due to broken links. 

We focus on the basic camera settings (aperture, shutter speed, and ISO) and program mode (auto or manual) because other tags, such as lighting, focal distance, white balance, are absent for most images. Overall, we found significant biases across all datasets. The distribution of camera settings is very long-tailed with distinct peaks at several F-number (2.8, 4, 5.6, 8), ISO (100, 200, 400), and shutter speed (1/125, 1/60, 1/30) values (see \cref{fig:ImageNet_camera_settings}). Since 50-80\% of all images were taken with auto settings (\cref{fig:datasets_exp_mode_dist}), we can infer that most of them are well-exposed. We can also group images with similar exposure by computing exposure values (EV) from camera parameters, as will be explained in \cref{sec:SNAP_dataset}. Distribution of the EVs across datasets shown in \cref{fig:datasets_EV_idx_dist} indicates that most images were taken in well-lit indoor and outdoor spaces. Finally, there is a significant chronological bias in vision data (\cref{fig:dataset_year_dist}); for instance, the bulk of images in ImageNet and COCO are from a decade ago and may no longer be representative of modern sensors and objects.

\subsection{SNAP dataset with balanced capture conditions}
\label{sec:SNAP_dataset}
To systematically investigate the effects of capture conditions, we collected a novel dataset, SNAP (for \textbf{S}hutter speed, ISO se\textbf{N}sitivity, and \textbf{AP}erture). It contains photos of multiple scenes taken under controlled lighting conditions and balanced across camera parameters. We designed SNAP for evaluating the models on 3 vision tasks: image classification, object detection, and visual question answering (VQA). This section describes SNAP's design, collection procedure, and properties.

\subsubsection{Data collection and properties}
\begin{wrapfigure}[16]{R}{0.45\textwidth}
\centering
\vspace{-2em}
\includegraphics[width=0.45\columnwidth]{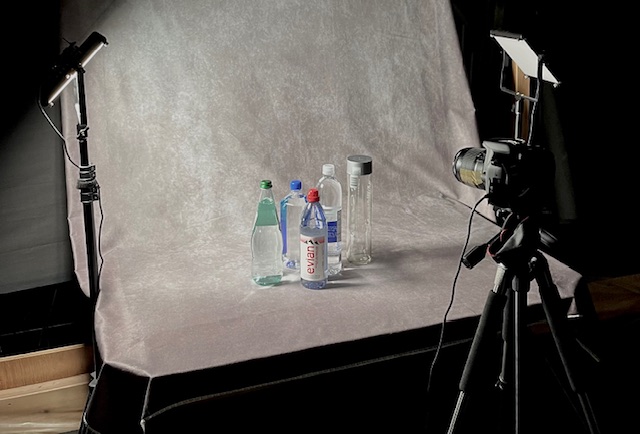}
\captionof{figure}{Data collection setup consists of a Canon EOS Rebel T7 camera on the tripod, a table covered with non-reflective cloth backdrop, and LED lights on either side.}
\label{fig:capture_setup}
\end{wrapfigure}

\noindent
\textbf{Capture setup.} \cref{fig:capture_setup} shows our data collection setup: a camera tethered to a laptop and placed on a tripod in front of a table covered with gray non-reflective cloth. Both the camera and the table are inside a blacked-out room where the only sources of light are 2 LED panels on either side of the table. To ensure consistent illumination, we use Yocto-Light-V3 lux meter \cite{yoctopuce}. All images are taken with Canon EOS Rebel T7 DSLR camera and saved as $1920\times1280$ JPEG files. 

On the camera side, we modify the shutter speed, F-Number, and ISO at 1-stop intervals within the available range of settings (see \cref{tab:canon_settings}). We use two illumination conditions, 1000 and 10 lux, roughly corresponding to an overcast day and twilight, respectively. Use of both high and low illumination allows expanding the range of usable camera settings. For example, under the low lighting conditions images can be taken with longer exposure times and higher ISO values. Other camera settings are fixed: flash is not used, focal length is set at 35 mm, and white balance is configured manually and fixed for each illumination condition. 

\noindent
\textbf{Object categories.} Dense sampling of camera parameters is time-consuming to collect. Thus to make the data suitable for all three vision tasks, we chose object categories present in both ImageNet and COCO---the most common datasets for training image classification and object detection/VQA models, respectively. SNAP contains the following 10 object categories that provide sufficient visual variety and physically fit within our setup: backpack, tie, water bottle, cup, laptop, mouse, remote, keyboard, cell phone, and comic book. Using only inanimate objects avoids motion artifacts during long exposure times and maintains layout constant throughout the capture. As our goal was to investigate the effects of camera settings, we minimized other effects: objects were photographed in canonical object poses (as models are sensitive to poses \cite{barbu2019objectnet}), with little occlusion and clutter, and against a neutral background (to prevent shortcuts \cite{geirhos2020shortcut}). 

\begin{wraptable}{R}{0.4\textwidth}
\centering
\vspace{-1.5em}
\caption{Canon EOS Rebel T7 camera settings used for data collection.}
\label{tab:canon_settings}
\resizebox{0.4\columnwidth}{!}{%
\begin{tabular}{@{}cl@{}}
\toprule
\textbf{\begin{tabular}[c]{@{}c@{}}Camera\\ parameters\end{tabular}} & \multicolumn{1}{c}{\textbf{Values}}  \\ \midrule
\begin{tabular}[c]{@{}c@{}}Shutter \\ speed (s)\end{tabular} &
  \begin{tabular}[c]{@{}l@{}}1/4000, 1/2000, 1/1000, 1/500, 1/250\\ 1/125, 1/60, 1/30, 1/15, 1/8, 1/4, 0.5,\\ 1, 2, 4, 8, 15, 30\end{tabular} \\ \midrule
ISO                                                                  & 100, 200, 400, 800, 1600, 3200, 6400 \\ \midrule
F-Number                                                             & 5.6, 8, 11, 16, 22                   \\ \bottomrule
\end{tabular}%
}
\vspace{-1em}
\end{wraptable}

In all images, there are 2 to 5 objects of the same class but with different appearance (e.g. a scene with 5 different water bottles in \cref{fig:capture_setup}). Having multiple objects in the scene adds minor occlusions, shadows, and clutter that in turn makes object detection and VQA tasks more realistic. Since all objects are of the same category, multi-label errors (common in the ImageNet \cite{stock2018convnets,tsipras2020imagenet,beyer2020we}) are avoided. 

\textbf{Capture procedure.}  For each scene and lighting condition, we first take one image with a camera's auto setting, then switch the camera to manual mode and use the gPhoto2 library \cite{gphoto2} to capture images over all possible combinations of F-number, shutter speed, and ISO sampled at 1-stop intervals. Photos with more than 95\% black (0) or white (255) pixels are discarded. One iteration through camera parameters for one lighting condition takes 40--60 minutes.

\textbf{Data composition and properties.} We captured 10 scenes with 2--5 objects from each object category (shown in \cref{fig:SNAP_overview}. There are a total of 37,558 images in the dataset, uniformly distributed across sensor settings. However, many combinations of shutter speed, aperture, and ISO lead to the same exposure, i.e., the same amount of light reaching the sensor (see examples in \cref{fig:exposure_equivalence}). This is known as exposure equivalence \cite[p.29]{prakel2009basics}. 
To identify such settings, we compute exposure value (EV) for each triplet using a standard formula \cite{ray2000camera}. Because exposure depends also on the lighting conditions, we group together settings with the same EV and lux values. We then re-index EV bins so that best exposure settings (i.e. closest to camera auto mode) receive a value of 0, over-exposed bins receive positive indices, and under-exposed are assigned negative ones. We refer to these indices as EV offset. For example, an EV offset of -1 means that the image is one stop underexposed, i.e., received half the light needed for the best exposure (see \cref{fig:ev_offset}). 

\subsubsection{Annotations and VLM prompts}
\label{sec:annotations}
All images in SNAP are provided with Exif information, object class label, bounding boxes, and segmentation masks. To test VLMs on image classification and object detection, we generated question-answer pairs from the annotations. Because the VLMs that we selected lack object detection abilities, we use counting (subitization) as a proxy since it requires localization. Due to the sensitivity of VLMs to prompting \cite{li2024can}, each question has open-ended (OE) and multi-choice (MC) variants:

\vspace{-0.5em}
\noindent
\textbf{Open-ended} \\
\textbf{Q1.} Objects of what class are in the image? Answer with the name of the class.\\
\textbf{Q2.} How many objects are in the image? Answer with one number.

\vspace{-0.5em}
\noindent
\textbf{Multiple choice} (answer options shuffled for each image) \\
\textbf{Q3.} Objects of what class are in the image? Select one of the following options: 10 classes + other \\
\textbf{Q4.} How many objects are in the image? Select one of the following options: A) 2; B) 3; C) 4; D) 5

\subsection{Models and metrics for fundamental vision tasks}
\label{sec:models_and_metrics}
We tested baseline and SOTA models, including 23 image classifiers, 16 object detectors, and 13 vision-language models (VLMs), on image classification, object detection, and visual question answering (VQA) tasks, respectively (see \cref{suppl:models}). We selected only open-source VLMs to be able to analyze their performance w.r.t. their training data and vision backbones. All selected models were tested on SNAP with default parameters, as well as pre- and post-processing routines where applicable. All experiments were performed on a cluster of 4 NVIDIA GeForce GTX 1080 Ti. The following metrics are used for evaluation:

\vspace{-1em}
\begin{itemize}[leftmargin=*]
\itemsep0em
\item{\textbf{Top-1 (\%)}}---a standard top-1 accuracy metric for image classification;
\item{\textbf{Localization-Recall-Precision (LRP)}} metric \cite{LRP_TPAMI, LRP_ECCV} for object detection. Unlike commonly used average precision (AP), LRP decouples classification and localization performance of the models. We report on the optimal LRP (oLRP), as well as localization accuracy (oLRP Loc), false positives (oLRP FP), and false negatives (oLRP FN). 
\item{\textbf{Soft and hard accuracy}} for the VQA task. Soft accuracy is simply \% of answers matching the ground truth. Hard accuracy is a stricter metric that penalizes partially correct answers. Following \cite{huang2025survey}, the answer is considered a factual error if it refers to something not present in the image and a faithfulness error if the answer does not adhere to the required format.
\item{\textbf{Parameter sensitivity (PS)}} measures sensitivity of the metrics to camera parameters on all tasks. As discussed in \cref{sec:SNAP_dataset}, SNAP contains many images of the same scenes that look nearly the same but are taken with different camera settings under different lighting levels (\cref{fig:exposure_equivalence}). To evaluate whether these barely perceptible changes affect model performance, we do the following: 1) we group the images of the same scene with the same EV offset; 2) to measure fluctuations in models' results within each set, we compute coefficient of variation (CV), defined as the ratio of the std to mean ($CV=\frac{\sigma}{\mu}$) of the metric; 3) we compute PS as the percentage of sets with $CV>1$. 
\end{itemize}

\subsection{Human experiment}
\label{sec:human_exp}
To establish a human baseline for vision performance under varying capture conditions, we use the VQA task described in \cref{sec:annotations} because it includes image classification (Q5), counting as a proxy for object detection (Q3), and allows direct comparisons to VLMs. We tested 43 human subjects on a subset of 8600 images from SNAP, evenly distributed w.r.t. object categories and capture conditions. The experiment was run in-person in the black-out room because online platforms (e.g. Amazon Turk) do not provide means to control lighting, display calibration, and presentation times, which are crucial for our study.  During the experiment, the subjects viewed the images on the computer screen and for each image, answered one of the categorization or counting questions listed in \cref{sec:annotations}. Since we intended to compare human answers to outputs of the predominantly feedforward DL vision models, the study was designed such that feedback processing in the brain was minimized. To achieve this, stimuli were shown very briefly (200 ms) and were immediately followed by the noise mask (200 ms), as in the past studies \cite{geirhos2018generalisation}. Full details are provided in \cref{suppl:human_experiment}.

\section{Experiment Results}

\subsection{Image classification}
\label{sec:image_classification}

\begin{wrapfigure}[18]{R}{0.65\textwidth}
\centering
\includegraphics[width=0.65\textwidth]{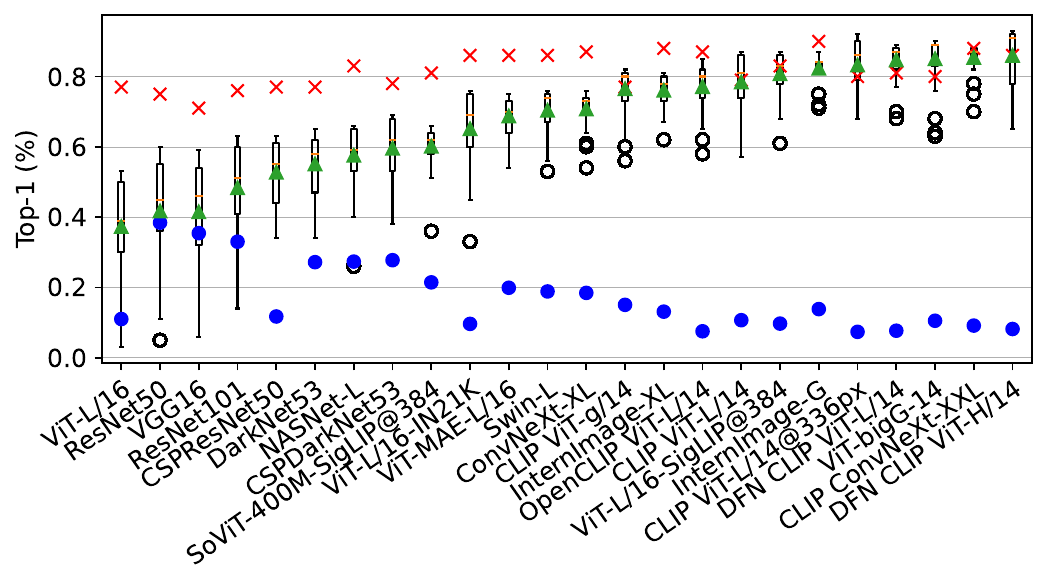}
\vspace{-1em}
\caption{Box plots show range and mean top-1 accuracy values for all models evaluated on SNAP. Blue circles show parameter sensitivity (PS) and red crosses mark top-1 accuracy on ImageNet.}
\label{fig:image_class_SNAP_vs_ImageNet}
\end{wrapfigure}

\textbf{Most models do not generalize to SNAP despite its simplicity and similarity to ImageNet.} Despite the intentional similarity of the SNAP objects to ImageNet and simplicity of the scenes, most models fail to generalize to our dataset. As shown in \cref{fig:image_class_SNAP_vs_ImageNet}, only 2 models (CLIP ViT-L/14@336px, DFN CLIP ViT-L/14) reach mean top-1 accuracy on SNAP comparable to their performance on ImageNet, while others show a significant decrease in accuracy ranging from 10 to 50\%.

\cref{fig:image_classification_data_size_vs_top1_acc} and \cref{fig:image_classification_model_size_vs_top1_acc} show plots of top-1 accuracy on SNAP relative to the model and training data size, respectively. Scaling the models and training data simultaneously improves the results; all models with performance above 80\% top-1 accuracy on SNAP have at least 200M parameters and are trained on 400M images or more.  The effects of the architecture and training are less pronounced in comparison. While 4 out of top-5 models are Transformers, some CNN-based architectures achieve competitive results (see \cref{fig:image_class_SNAP_vs_ImageNet}). For example, supervised training of vanilla ViT L/16 and VGG-16 on ImageNet results in similar performance and so does CLIP-pretraining of ViT and ConvNeXt on comparable amounts of data.  

\begin{wrapfigure}[23]{R}{0.6\textwidth}
\centering
\vspace{-1.5em}
\includegraphics[width=0.6\textwidth]{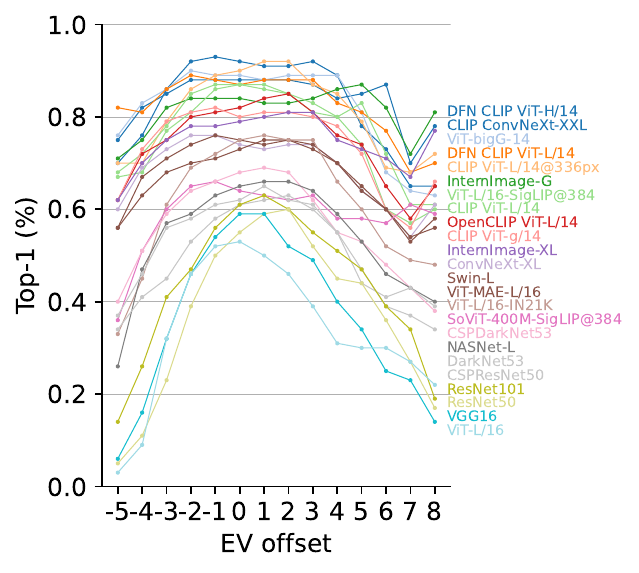}
\captionof{figure}{Image classification top-1 (\%) across exposure levels in SNAP. Each dot represents mean accuracy across all images in the corresponding EV offset bin.}
\label{fig:img_class_EV}
\end{wrapfigure}

The models trained or fine-tuned on ImageNet struggle with some categories in SNAP. For instance, most models misclassify cups and laptops as mugs and notebooks, respectively, due to inconsistent labeling in ImageNet. Hence, in evaluation, we accept both answers as correct. Most models also underperform on phones, which is the consequence of the chronological bias pointed out earlier in \cref{sec:dataset_analysis}. Because most images in ImageNet were taken prior to 2009, it is dominated by photos of outdated phone designs. Only the zero-shot CLIP-pretrained models (not finetuned on ImageNet) were able to classify modern phones in SNAP correctly.

\noindent
\textbf{Models reach peak top-1 accuracy on well-exposed images but perform significantly worse on under- and over-exposed images.} 
Exposure, i.e. the combined effect of camera settings and illumination, affect all models (see \cref{fig:img_class_EV}). The models reach their peak performance on well-exposed images (EV offset between -2 and 2). 8 out of top-10 models match or surpass their ImageNet results on this subset of SNAP, which is expected given the intentional similarity of the objects and scenes. At the same time, all models perform worse on over- and under-exposed images. This performance drop is not symmetric; accuracy on the under-exposed images is lower than on the over-exposed images.

Again, the larger models (regardless of the architecture) trained on more data achieve higher peak accuracy on well-exposed images and are less affected even by the most extreme exposures. For instance, CLIP-pretrained ConvNeXt-XXL reaches over 70\% accuracy on nearly all-black (EV offset -5) and all-white (EV offset 8) images. Notably, this model is trained on the 2B subset of LAION with minimal data augmentations (only random cropping and erasing). This agrees with our analysis in \cref{suppl:cv_datasets} that shows the higher diversity of manual settings and exposures in LAION.

\textbf{All models are susceptible to image perturbations caused by slight variations of camera parameters.} This is evident from high parameter sensitivity (PS) w.r.t. top-1 accuracy that reaches over 20\% some models (\cref{fig:img_class_EV}). Even the top-5 performing models inconsistently classify nearly 10\% of the scenes that look essentially the same. High sensitivity of all models is apparent even on well-exposed images, which better match the properties of the training data.

\subsection{Object detection}

\begin{wrapfigure}[15]{R}{0.6\textwidth}
\vspace{-3em}
\centering
\includegraphics[width=0.6\textwidth]{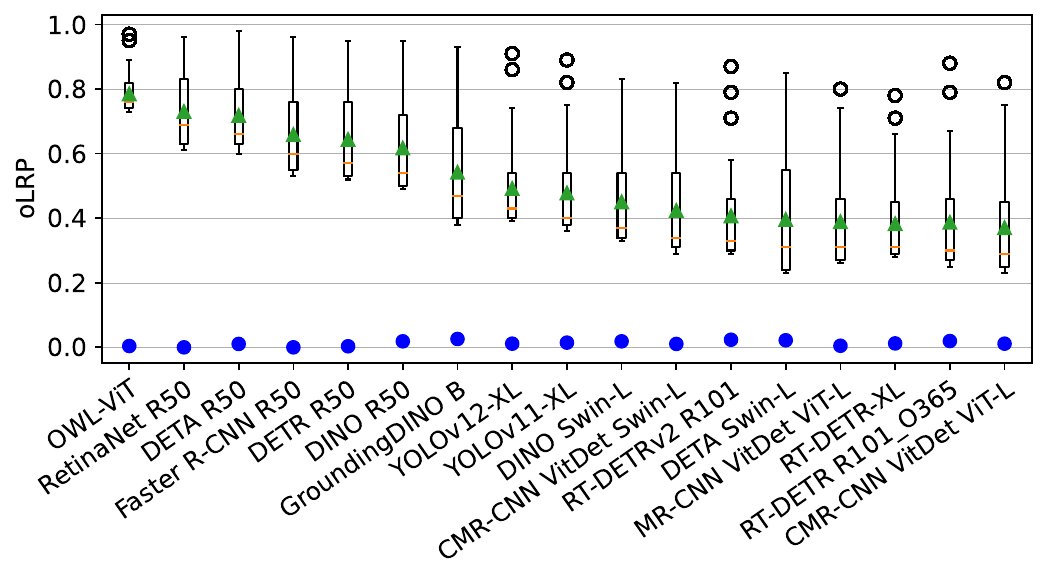}
\caption{Box plots show range and mean oLRP values for all object detection models evaluated on SNAP. Blue circles show parameter sensitivity (PS) w.r.t. oLRP.}
\label{fig:obj_det_EV_offset_oLRP_box}
\end{wrapfigure}

\textbf{Many models perform well overall, but SNAP is simpler than most object detection datasets.} \cref{fig:obj_det_EV_offset_oLRP_box} shows the results of testing object detection models on the SNAP dataset. For this experiment, we also computed the mean AP scores of the models to compare against the COCO benchmark results. The mean AP of the models is approx. 15-20\% higher on COCO (\cref{fig:object_detection_EV_offset_AP_box}). This can be explained by the relative simplicity of SNAP: the scenes are not cluttered, objects appear large against plain background, and there is little occlusion or cropping (see \cref{fig:SNAP_overview}). In comparison, the scenes in COCO are busy, filled with multiple small objects, many of which are only partially visible due to occlusion or cropping.

\begin{wrapfigure}[21]{R}{0.5\textwidth}
\centering
\vspace{-1.5em}
\includegraphics[width=0.5\columnwidth]{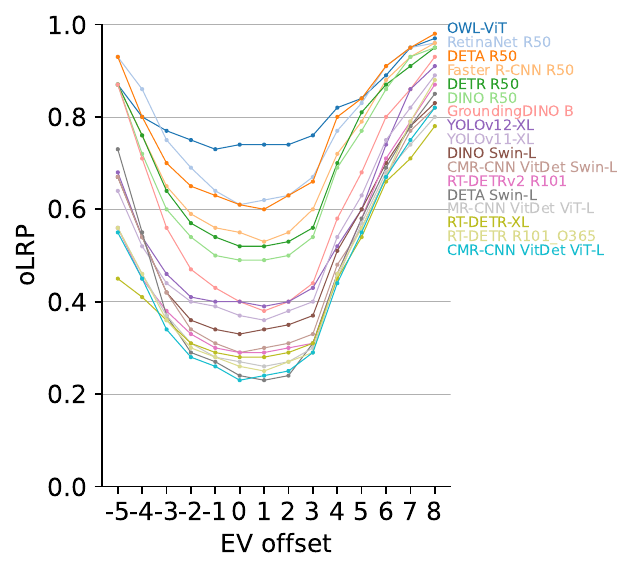}
\captionof{figure}{Object detection at different exposure levels in SNAP. Each point is a mean oLRP of the model predictions on a given EV offset. Model labels are sorted by oLRP from worst to best)}
\label{fig:obj_det_EV_offset_oLRP_line}
\end{wrapfigure}

The majority of existing object detection models are pre-trained on ImageNet and COCO, therefore the effects of scaling data cannot be explored fully. However, similar to image classification, there is a correlation between model scale and performance. For instance, the top model DETA Swin-L is trained on the Object365 dataset \cite{shao2019objects365} with 600K images and 10M bounding boxes in addition to COCO. At the same time, RT-DETR is trained only on COCO and performs on par.

\noindent
\textbf{Effects of exposure on object detection are more pronounced than on image classification.} Similar to image classifiers, the overall performance (oLRP) of object detectors degrades on under- and over-exposed images, as illustrated in \cref{fig:obj_det_EV_offset_oLRP_line}. In this case, however, the drop-off towards either extreme is more pronounced and is apparent even in the well-exposed range for all models. There is also asymmetry in the oLRP scores, but the trend is reversed---unlike image classifiers, the object detection models perform worse on the over-exposed images. 

Exposure affects both localization and classification, but to a different extent. Localization (oLRP Loc) and false positive (oLRP FP) errors increase for under- and over-exposed images, but remain relatively low for all models (see \cref{fig:obj_det_EV_offset_oLRP_FP_line} and \cref{fig:obj_det_EV_offset_oLRP_Loc_line}). Misclassifications (oLRP FN) are the largest contributor to the overall oLRP score. They mirror the pattern of top-1 accuracy on the image classification task: errors are fairly low for well-exposed images but rise quickly toward larger EV-offsets in both directions (as shown in \cref{fig:obj_det_EV_offset_oLRP_FN_line}). The majority of object detectors use ImageNet pre-trained backbones (typically, variants of ResNet or ViT) and are further trained only on COCO for the object detection task. Therefore, misclassification errors are likely an artifact of ImageNet pre-training. Several exceptions from this trend lend further support to this conclusion. Three models (DETA Swin-L, DINO Swin-L, and Grounding DINO) with consistently low oLRP Loc and oLRP FP scores across all EV offsets are all trained on ImageNet21K and O365. Similarly, OWL-ViT, which uses a large CLIP ViT backbone and is trained on O365 \cite{shao2019objects365} and VG \cite{krishna2017visual} for detection, has a flatter curve (albeit at a higher error), reminiscent of CLIP ViTs in \cref{fig:img_class_EV}.

\noindent
\textbf{Most of the fluctuations on images with similar exposure come from misclassifications (FN errors), whereas localization is less affected.} While the overall sensitivity of the oLRP scores for all models is low, localization and classification performance is still affected. For nearly all models, SP is highest for the oLRP FN component, reaching nearly 20\% for some.

\subsection{Visual question answering}
\label{sec:vqa}

\begin{wrapfigure}[14]{R}{0.5\textwidth}
\centering
\vspace{-2em}
\includegraphics[width=0.5\textwidth]{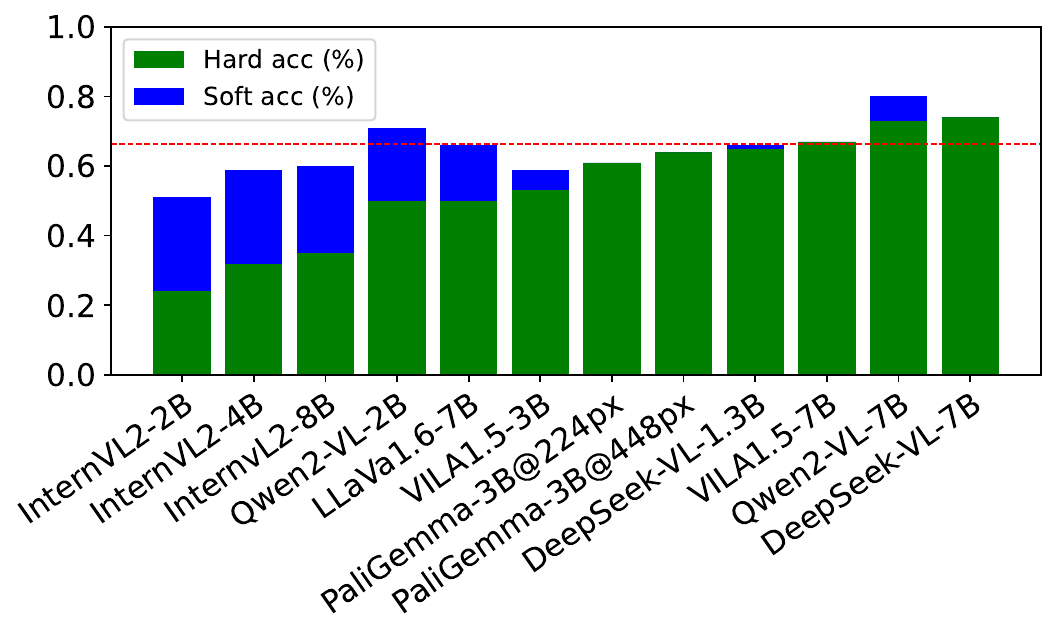}
\vspace{-1em}
\captionof{figure}{Mean hard accuracy (green) and soft accuracy (blue) on all questions for models and human subject accuracy (red line).}
\label{fig:vqa_acc_vs_acc1}
\end{wrapfigure}

\textbf{VLMs are comparable to humans in average accuracy across all questions.} We first assess the average performance of the tested VLMs against the human subjects by aggregating accuracy across all images and questions. Computing the accuracy scores for VLMs is a challenge. In many cases, matching the answer of the models with the ground truth may inflate the scores because not all models follow the prompt well. We used a combination of simple regular expressions and manual clean-up to bring all model answers to the same format (see \cref{suppl:evaluation}).

Soft and hard accuracy scores in \cref{fig:vqa_acc_vs_acc1} show percentage of partially correct and correct answers. The majority of the partially correct answers are due to not following the prompt. Although the questions explicitly ask to answer with a single number or multi-choice option (see \cref{sec:annotations}), many VLMs instead return multiple categories of objects or generate a long explanation. 
\begin{wrapfigure}[16]{R}{0.6\textwidth}
\centering
\vspace{-1.5em}
\includegraphics[width=0.6\columnwidth]{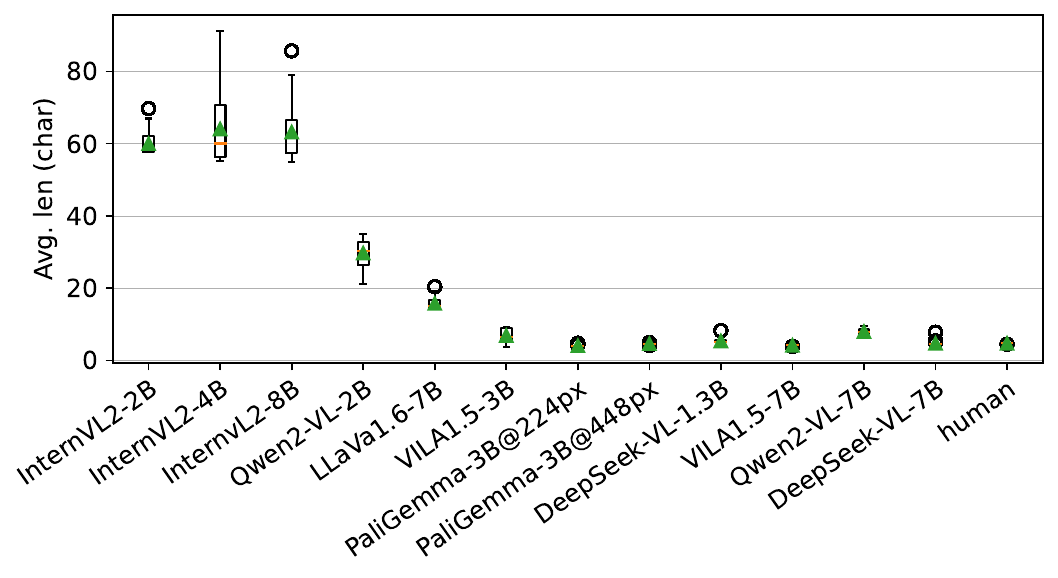}
\captionof{figure}{Mean length of responses (number of characters) across all questions. Longer answers indicate presence of faithfulness errors.}
\label{fig:vqa_EV_offset_avg_len_box}
\end{wrapfigure}
As we expect answers not exceeding 10-15 characters for most questions, the length of the answer serves as a proxy for these types of mistakes and shows varying ability of the models to follow the prompt in \cref{fig:vqa_EV_offset_avg_len_box}. For example, PaliGemma, VILA, and DeepSeekVL-7B almost never deviate from the answer format, as do human subjects, whereas all InternVL models and Qwen2-VL-2B tend to produce lengthy justifications and unprompted chain of thought explanations. Both hallucinated non-existing answer options and even switched to a different language. Overall, there is a trend towards fewer hallucinations in models that have larger LLM component. For instance, Qwen2-VL-7B hallucinates substantially less compared to its 2B-parameter version. Similarly, InternVL2-8B is not as prone to hallucinations as 4B and 2B models from the same family.

Discounting partially correct answers, 4 models reach or surpass the average accuracy of human subjects (66\%): Qwen2-VL-7B-Instruct (77\%), DeepSeek-VL-7B-chat (74\%), VILA1.5-7b (68\%), and DeepSeek-VL-1.3B-chat (66\%). All three use the largest vision models trained on the most data. DeepSeek and VILA use ViT-L-16-SigLIP-384 trained on 100B image pairs from the proprietary Webli dataset \cite{chen2023pali}, whereas Qwen2-VL uses DFN5B-CLIP-ViT-H-14 vision encoder, which ranked first on the image classification task in \cref{sec:image_classification}. An interesting exception is PaliGemma: on one hand, its hallucination rate is near zero, on the other, its vision backbone, a shape optimized SigLIP ViT (SoViT-400M-SigLIP) pre-trained on Webli, is weaker than others. On SNAP, it achieves only 60\% top-1 accuracy, compared to the 81\% of the SigLIP-ViT variant used by DeepSeek and VILA.

\textbf{VLMs are less affected by under-exposed images than humans but do not reach human peak performance on well-exposed images.} We also looked at the performance of the models at different exposure levels. 
As in other tasks, performance of the models reaches its peak on the well-exposed images and drops off outside of that range. Despite lower overall results, human subjects reached a peak mean accuracy of 89\% across all questions on well-exposed images (EV offset of 1), whereas the top-3 VLMs mentioned above were at 80\%, 78\% and 77\%, respectively. Only on the MC categorization question, 4 top VLMs outperformed peak human accuracy (98.3\%) on well-exposed images. 
\begin{wrapfigure}[20]{r}{0.5\textwidth}
\centering
\vspace{-1em}
\includegraphics[width=0.5\columnwidth]{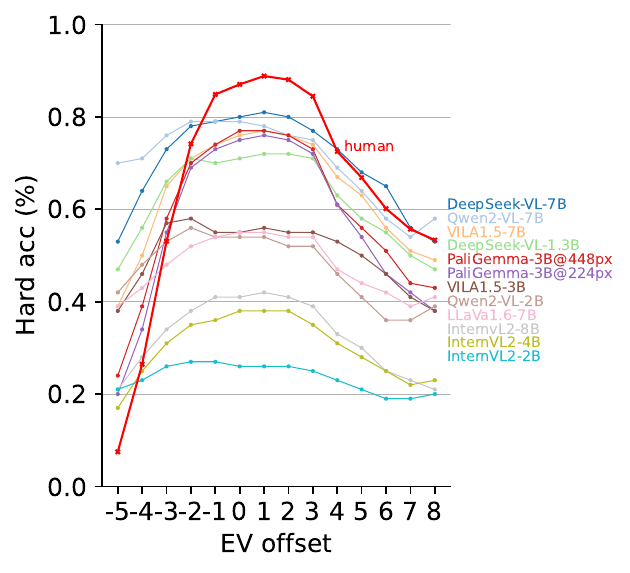}
\captionof{figure}{Mean hard accuracy on all questions and exposures in SNAP for VLMs and humans.}
\label{fig:vqa_EV_offset_acc1_line}
\end{wrapfigure}
The low average performance of the human subjects is mainly due to sharp drop-off on under-exposed images (EV offset < -3). These images appear mostly black so the performance of the human subjects reduces to chance. On open-ended questions, many subjects explicitly indicated not being able to answer. While some models do that too (e.g. PaliGemma responded "unanswerable" in such cases), they do so inconsistently. On the over-exposed images, however, the human subjects maintained high level of performance surpassing nearly all VLMs. Exposure affects the rate of hallucinations as well, whereas the human responses remain the same throughout. Most models tend to produce longer answers outside the well-exposed range. The only exceptions are the Qwen models that do the opposite.

\textbf{All VLMs are sensitive to camera parameter variations on all questions but performance changes are unpredictable.} Likely due to the interactions between the vision and language components, each model had different performance trends that varied significantly depending on the question type (categorization and counting) and options (multi-choice or open-ended), as shown in \cref{fig:vqa_all_box_acc1}. For example, the strongest overall model DeepSeek-VL-7B is very consistent on MC categorization (Q5), but reaches 10-20\% PS on other questions. The second best model Qwen2-VL-7B has the lowest overall PS on OE categorization (Q3) compared to other models, while its PS on other questions is relatively high (15-20\%).  

\section{Conclusions}

We conducted a systematic study to determine the sensitivity of DL vision algorithms to capture bias that includes changes in both camera parameters and illumination. We first identified a significant bias in the training data---even datasets with millions of images are highly imbalanced in terms of camera settings. We then constructed a novel dataset SNAP with more uniform capture conditions. This dataset was used to test a number of models and human subjects on common vision tasks. Three main conclusions can be made from this study. First, despite the intentional simplicity of SNAP, most models do not match their performance on other benchmarks and few surpass average performance of humans. This gap persists even for the largest models that are trained on billion-scale data. Second, both models and humans are sensitive to deviations of exposure, but in different ways. This discrepancy and the fact that most models do not match human performance even in the optimal cases again points to the generalization issues. In addition, our analysis shows that pre-trained image classification backbones propagate their capture biases on downstream tasks. Third, all models are sensitive even to the slight variations in camera settings on all tasks. This is especially concerning for practical applications, where precise control of camera or illumination conditions is not possible. 

\textbf{Limitations.} Due to the time-consuming process of gathering real-world data, we were not able to test other aspects of capture bias, such as effects of sensor type. We used Canon DSLR for data collection primarily because it offered better settings control, however, it is also a very common camera make in the vision datasets. Mobile phones and webcams are less represented and are more difficult to control but would be more relevant for practical applications. Canon DSLR did not allow exploring the full range of settings found in the datasets, particularly for aperture. Furthermore, sensor settings were sampled at 1-stop intervals, rather than 1/2 or 1/3 stops, and with only 2 lighting conditions because otherwise capture would take prohibitively long (multiple hours per scene).

{\small
\bibliographystyle{plain}
\bibliography{psychosal}
}

\newpage
\appendix
\counterwithin{figure}{section}
\counterwithin{table}{section}
\setcounter{figure}{0}
\setcounter{table}{0}

\section{Computer vision dataset properties}
\label{suppl:cv_datasets}

\begin{multicols}{2}
   \centering
   \includegraphics[width=\columnwidth]{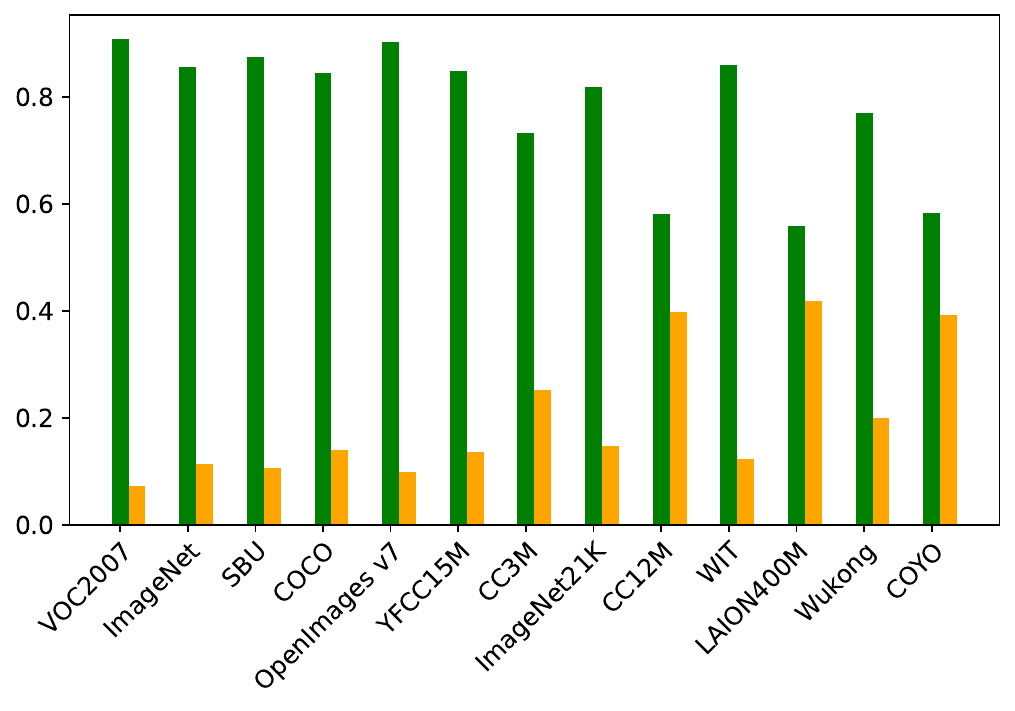}
   \captionof{figure}{Distribution of images in the datasets taken with auto or manual camera settings.}
   \label{fig:datasets_exp_mode_dist}
   \columnbreak
   \includegraphics[width=\columnwidth]{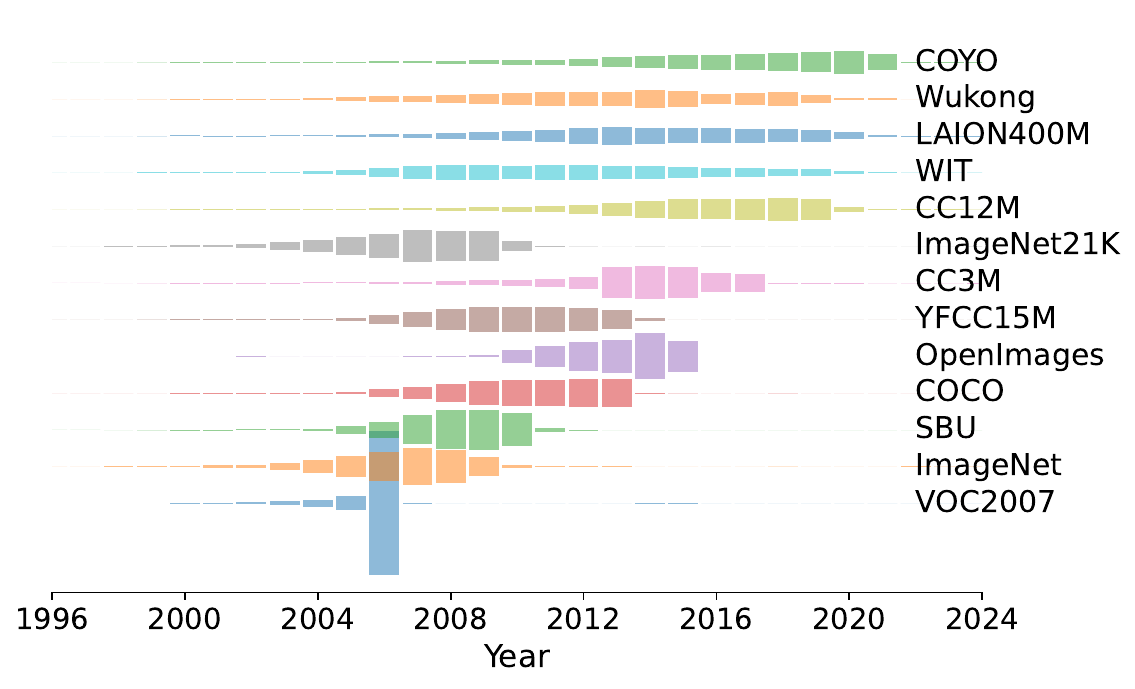}
   \captionof{figure}{Distribution of images in the datasets by year the images were taken. Each bar plot corresponds to a single dataset. Plots are arranged in chronological order from bottom to top.}
  \label{fig:dataset_year_dist}
\end{multicols}

In this section we provide additional details on the properties of the common computer vision dataset. For this analysis we selected 13 common datasets that are used to train models for image classification, object detection, and fundamental vision models. The largest amount of data for our analysis comes from the latter category, namely datasets that contain image-text pairs collected from the Internet. We excluded several datasets, such as Object365 \cite{shao2019objects365} and VisualGenome \cite{krishna2017visual}, because all image metadata was stripped and sources for the images (e.g., URL links or Flickr ids) were not provided.

\cref{fig:datasets_exp_mode_dist} shows the distribution of images taken with automatic and manual camera settings. To identify mode, we check \textit{ExposureMode} tags in image Exif data. The auto mode category includes various settings and presets that either fully or partially automate exposure. The most common tag values include: ``Auto'', 'Auto exposure', 'Aperture-priority AE', 'Auto bracket', 'Creative (Slow speed)', 'Shutter speed priority AE', 'Landscape', 'Portrait','Action (High speed)','Normal program', etc. Tag values corresponding to the manual mode are: ``Manual'' and `Manual exposure''. Overall, auto exposure was used to take 72\% of all photos (with Exif data) in all datasets we analyzed and 26\% of data was captured in manual mode.

\begin{figure}[t!]
\centering
\includegraphics[width=\textwidth]{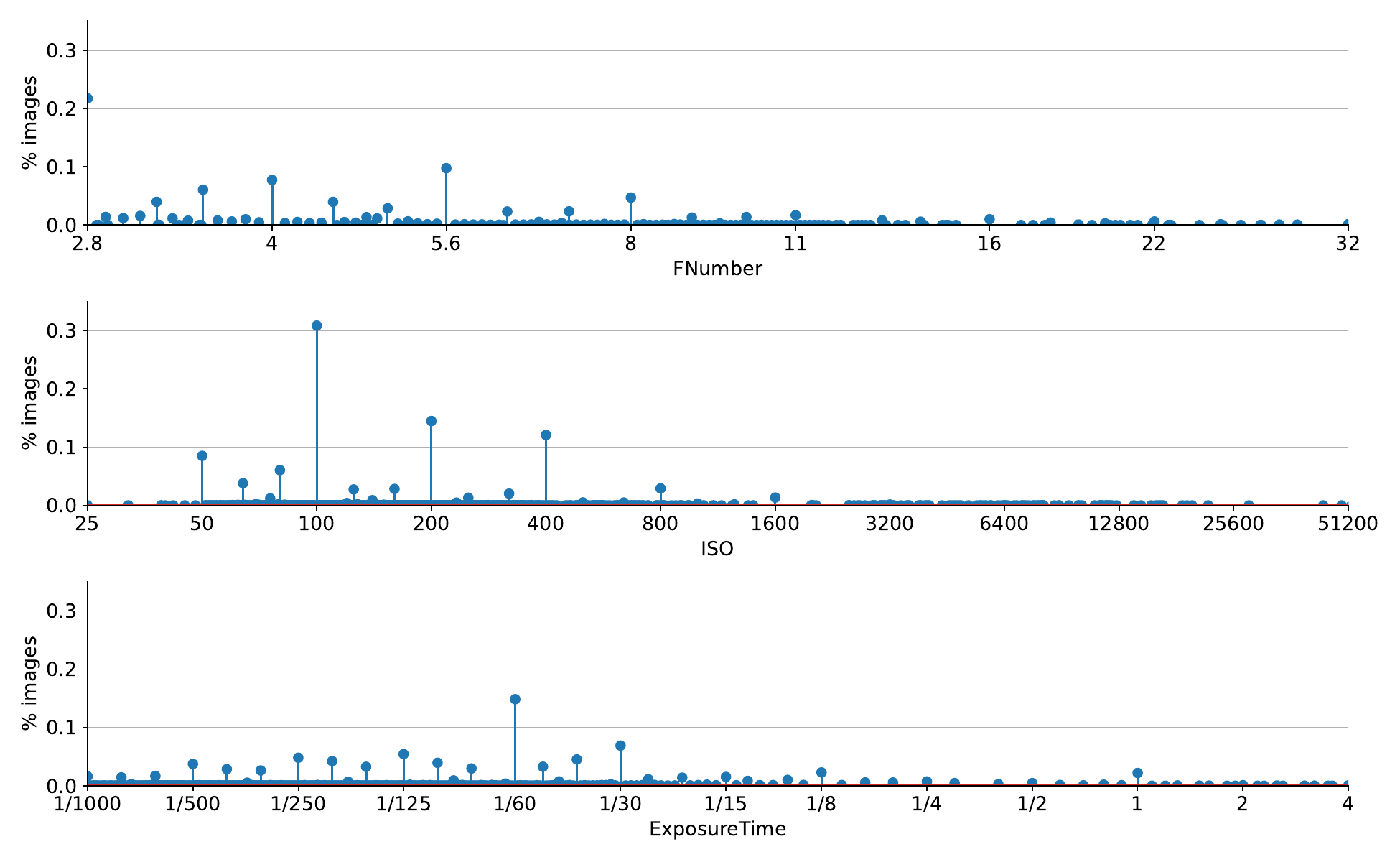}
\caption{Normalized distribution of camera parameter settings (F-Number, ISO, and exposure time) in the ImageNet dataset. Horizontal axis is plotted on a log-scale.}
\label{fig:ImageNet_camera_settings}
\end{figure}

We also looked at the distribution of values of 3 camera parameter values in the datasets. \cref{fig:ImageNet_camera_settings} shows an example from the ImageNet, which is characteristic of the other datasets as all of them are very long-tailed and have distinct peaks at 1-stop intervals in all 3 settings.

\begin{figure}[h!]
\centering
\includegraphics[width=0.23\columnwidth]{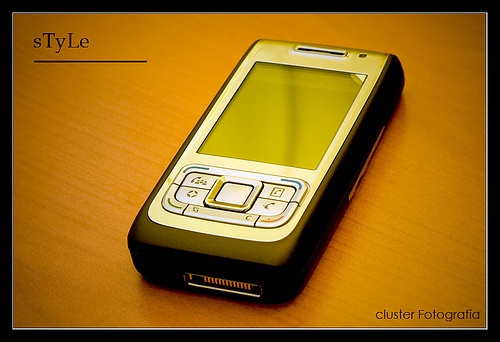}
\includegraphics[width=0.23\columnwidth]{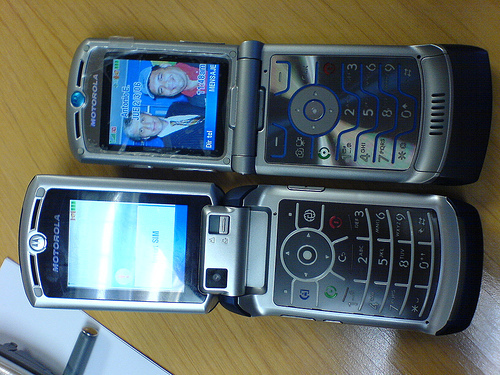}
\hspace{1em}
\includegraphics[width=0.23\columnwidth]{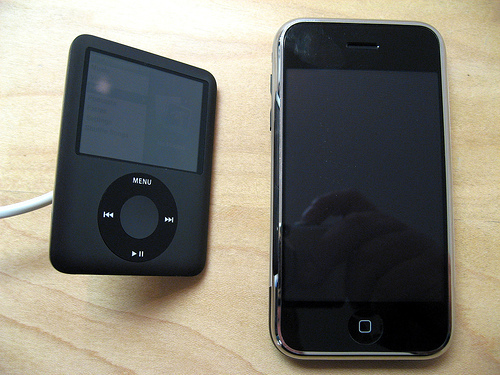}
\includegraphics[width=0.23\columnwidth]{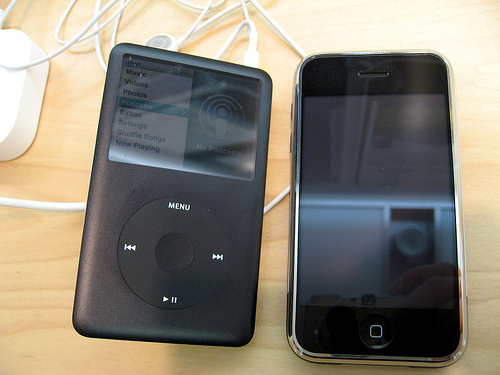}
\caption{Examples of chronological bias in ImageNet: images labeled as ``phone'' are on the left and ``iPod'' on the right.}
\label{fig:ImageNet_chrono_bias}
\end{figure}

Lastly, many models trained or fine-tuned on ImageNet had a much lower performance on the phone class (<30\% top-1 accuracy). Even one of the top models CLIP ConvNeXt-XXL reached only 53\% accuracy on this class. All models frequently misclassified phones as iPods. This issue likely stems from the chronological bias of ImageNet, where most images were taken more than a decade ago. \cref{fig:ImageNet_chrono_bias} shows examples from the \textit{phone} and \textit{iPod} synsets. Images for the phone synset are dominated by flip-phones and keyboard phones, whereas iPod class often features iPhones that look similar to the modern phone designs. 

\section{SNAP dataset properties}

\begin{figure}[t!]
\centering
\includegraphics[width=\columnwidth]{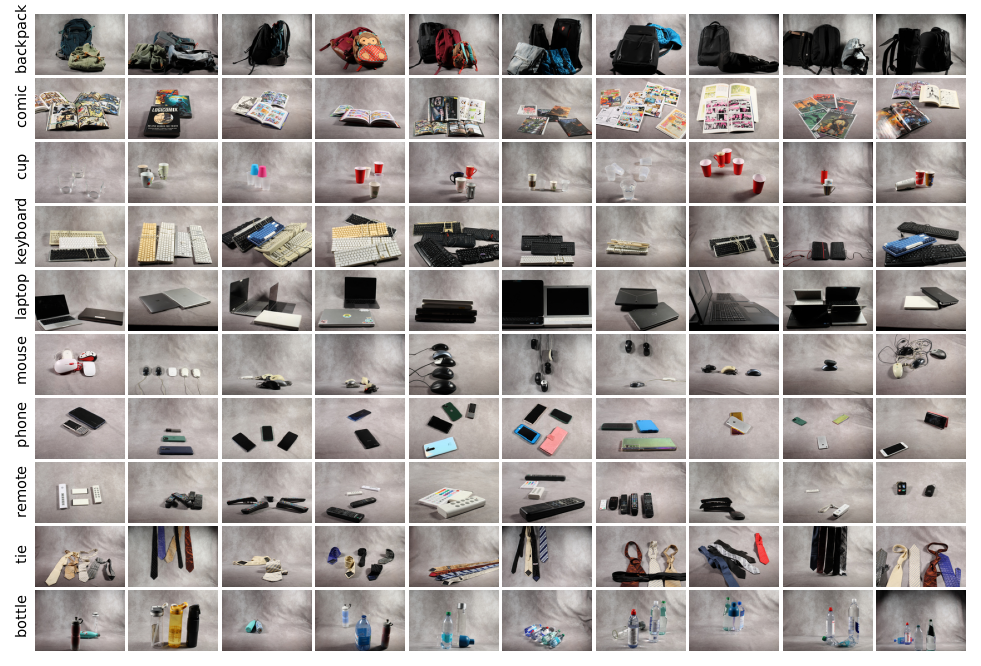}
\caption{Overview of the SNAP dataset. For each of the 10 object categories, we capture 10 scenes with different number of objects in different configurations. Each of these scenes is captured with the full range of camera parameters under 2 illumination conditions.}
\label{fig:SNAP_overview}
\end{figure}

Here, we provide additional details and qualitative samples from the proposed SNAP dataset. Figure \cref{fig:SNAP_overview} shows a general overview of the 100 unique scenes (10 for each of the 10 objects) captured in the dataset. Each of the scenes was photographed with the full range of Canon EOS Rebel T7 camera parameters (listed in \cref{tab:canon_settings}) under 2 illumination conditions.

\begin{figure}[t!]
\centering
\includegraphics[width=0.7\columnwidth]{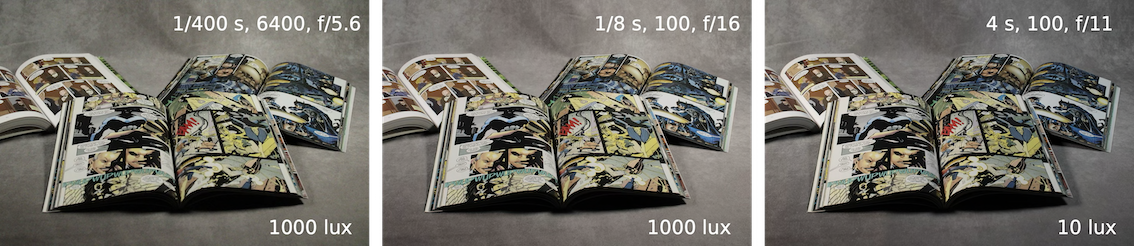}
\caption{An illustration of the exposure equivalence. All photos are taken with the same exposure but this is achieved with 4 different camera parameters (listed in the top-right corner) and under different illumination conditions.}
\label{fig:exposure_equivalence}
\end{figure}

Due to exposure equivalence, many photos of the same scene look very similar, despite being taken with different camera parameters and under different level of lighting. This is illustrated in \cref{fig:exposure_equivalence}. To find sets of images with the same camera settings, we estimate exposure value (EV) index from camera parameters using a standard formula from \cite{ray2000camera}: $\mathrm{EV} = \mathrm{log}_2(\frac{\mathrm{F-Number}^2}{\mathrm{ExposureTime}}) - \mathrm{log}_2(\frac{\mathrm{ISO}}{100})$. These EV values were used to produce plots in \cref{fig:datasets_EV_idx_dist}.

\begin{figure}[t!]
\centering
\includegraphics[width=0.8\columnwidth]{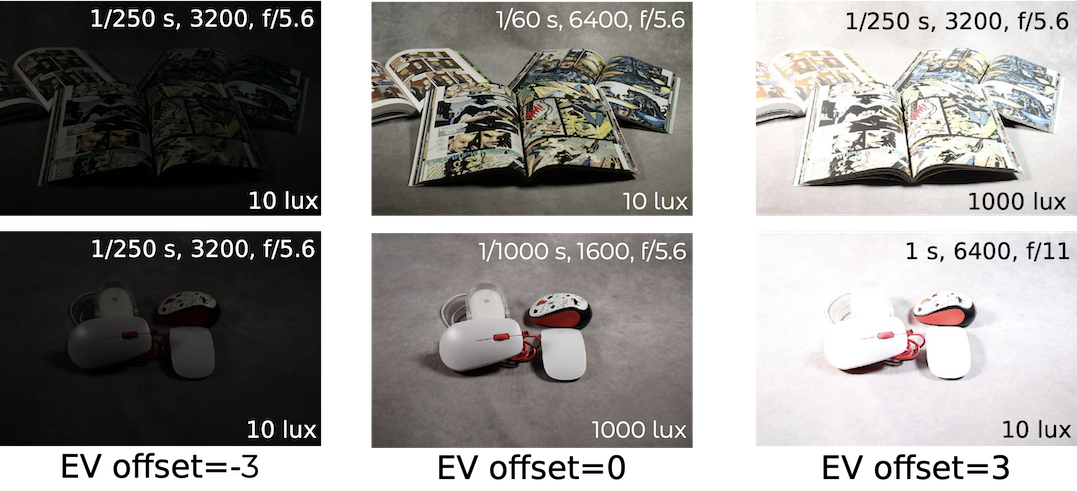}
\caption{Illustration of the EV offsets corresponding to different exposure levels. Bin with the ``optimal'' exposure (closest to the auto settings) is assigned an EV offset of 0 and the remaining bins are indexed relative to it. Negative and positive EV offsets correspond to under-exposed and over-exposed images, respectively.}
\label{fig:ev_offset}
\end{figure}

Exposure depends also on the illumination level, which is not reflected in this formula. Therefore, images with the same EV value but taken in different lighting conditions will look different. To bin together images that have the same exposure, regardless of illumination, we introduce EV offset (similar to exposure compensation). For each illumination condition, we find EV index that is the closest to the auto settings and assign EV offset of 0 to that bin. We compute the other EV offsets from EV relative to the 0 bin. For the 1000 lux condition we use EV of 11 as a starting point EV of 5 for the 10 lux. Now, all images with EV offset of 0 are well-exposed and those with negative or positive EV offsets are under- or over-exposed, respectively. Because camera settings in SNAP were sampled at 1-stop intervals, increase/decrease by 1 EV offset means doubling/halving of the amount of light reaching the sensor. An illustration of images with different EV offsets is provided in \cref{fig:ev_offset}.

\section{Human experiment}
\label{suppl:human_experiment}

To establish a human baseline for the categorization and subitization task under different capture conditions, we tested 43 subjects (25 M, 17 F, 1 did not specify), ranging in age from 20 to 67 years old (Mean=30.2, SD=10.5), on a subset of images from the SNAP dataset. Subjects were recruited through the university mailing list and posters on the bulletin boards on campus. The study posed minimal risk to participants and received ethics approval from the IRB (certificate \# e2025-053). All subjects signed consent forms and were compensated 20CAD for their participation in the experiment, which took approximately 30 minutes. 

Because our dataset consists of images with different levels of exposure, controlling display calibration and lighting conditions is crucial as it can affect what the subjects see on the screen. Thus, we conducted the in the dark room in our lab. Subjects sat 60 cm away from the calibrated 22" LCD monitor and viewed images (shown at 960$\times$640 px resolution) on the screen. All subjects had normal or corrected-to-normal vision.

We tested the subjects on the same set of questions as the VLMs. Recalling \cref{sec:annotations}, there were 4 questions in total: subitization (how many objects are in the image?) and categorization (what class of objects are in the image?), each with 2 versions---multiple-choice (MC) and open-ended (OE). 

Subjects were randomly assigned one of two groups: group 1 answered MC subitization and OE categorization questions, group 2 answered OE subitization and MC categorization questions. This was done to prevent priming effects, i.e. subjects that had answered the MC categorization question did not answer the OE version of the same question as they would be familiar with the available options.

Each subject performed 200 trials, 100 for each question, split into blocks of 50. Since we have 100 unique scenes in the SNAP dataset, each scene was seen twice. To minimize the chance of subjects recognizing previously seen images, we sampled pairs of images at least 4 stops apart. Furthermore, both the order of blocks and order of trials within each block were randomized. At the start of each block, the question appeared on the screen and remained the same for the all images in the block. No feedback was given to the participants during the experiment. The same questions and answer options for MC questions were used in the experiment and to test VLMs to allow direct comparisons. 

Each trial proceeded as follows: 1) fixation cross was shown for 1s; 2) image appeared for 200 ms followed by a white mask for 200 ms; 3) a question prompt appeared: for open-ended question, a text box was provided where the subject could enter the answer using the keyboard; for multiple choice questions, options were shown on the screen to be selected via mouse; 4) after answering the question, subjects pressed space to continue to the next trial. The experiment was not timed and only the answers were recorded. We used PsychoPy \cite{peirce2019psychopy2} to control the experiment.

A practice session consisting of 20 trials (10 for each question) was conducted prior to the experiment to familiarize the subjects with the procedure. This session was not recorded and used images from a separate set with different object categories (tv, umbrella, spoon, orange, banana, sports ball, potted plant).

The subjects performed a total of 8600 trials with unique images from SNAP, evenly distributed w.r.t. to object categories and capture conditions.

\section{Models}
\label{suppl:models}

Below we list baseline and SOTA models, including 23 image classifiers (\cref{tab:img_class}), 16 object detectors (\cref{tab:obj_det}), and 13 vision-language models (\cref{tab:vlm}), for image classification, object detection, and visual question answering (VQA) tasks, respectively. 

\begin{table}[]
\centering
\caption{Image classification models used for experiments. Model name is how the model is referred to in the text and figures. In addition, we provide the original model id and URL.}
\label{tab:img_class}
\resizebox{0.8\columnwidth}{!}{%
\begin{tabular}{@{}ll@{}}
\toprule
Model name                                   & Model id, URL                                                                                               \\ \midrule
ViT-L/16-SigLIP@384 \cite{2023_ICCV_Zhai}    & \href{https://huggingface.co/timm/ViT-L-16-SigLIP-384}{ViT-L-16-SigLIP-384}                              \\
SoViT-400M-SigLIP@384 \cite{2023_ICCV_Zhai}  & \href{https://huggingface.co/google/siglip-so400m-patch14-384}{siglip-so400m-patch14-384}                \\
OpenCLIP ViT-L/14 \cite{2023_CVPR_Cherti} &
  \href{https://huggingface.co/timm/vit_large_patch14_clip_224.laion2b_ft_in12k_in1k}{vit{\_}large{\_}patch14{\_}clip{\_}224.laion2b{\_}ft{\_}in1k} \\
DFN CLIP ViT-H/14 \cite{2023_arXiv_Fang_DFN} & \href{https://huggingface.co/apple/DFN5B-CLIP-ViT-H-14-378}{ViT-H-14-quickgelu{\_}dfn5b}                 \\
DFN CLIP ViT-L/14 \cite{2023_arXiv_Fang_DFN} & \href{https://huggingface.co/apple/DFN2B-CLIP-ViT-L-14}{ViT-L-14-quickgelu{\_}dfn2b}                     \\
ViT-MAE-L/16 \cite{2022_CVPR_He_ViT_MAE}     & \href{https://github.com/facebookresearch/mae/blob/main/FINETUNE.md}{vit{\_}large{\_}patch16{\_}mae}     \\
CLIP ConvNeXt-XXL \cite{2022_CVPR_ConvNeXt} &
  \href{https://huggingface.co/timm/convnext_xxlarge.clip_laion2b_soup_ft_in1k}{convnext{\_}xxlarge.clip{\_}laion2b{\_}soup{\_}ft{\_}in1k} \\
ConvNeXt-XL \cite{2022_CVPR_ConvNeXt} &
  \href{https://huggingface.co/timm/convnext_xlarge.fb_in22k_ft_in1k_384}{convnext{\_}xlarge.fb{\_}in22k{\_}ft{\_}in1k} \\
ViT-bigG-14 \cite{2021_ICML_Radford} &
  \href{https://huggingface.co/laion/CLIP-ViT-bigG-14-laion2B-39B-b160k}{ViT-bigG-14{\_}laion2b{\_}s39b{\_}b160k} \\
CLIP ViT-L/14@336px \cite{2021_ICML_Radford} & \href{https://huggingface.co/openai/clip-vit-large-patch14-336}{clip-vit-large-patch14-336}              \\
CLIP ViT-L/14 \cite{2021_ICML_Radford}       & \href{https://huggingface.co/openai/clip-vit-large-patch14}{clip-vit-large-patch14}                      \\
CLIP ViT-g/14 \cite{2021_ICML_Radford}       & \href{https://huggingface.co/laion/CLIP-ViT-g-14-laion2B-s12B-b42K}{ViT-g-14{\_}laion2b{\_}s12b{\_}b42k} \\
ViT-L/16-IN21K \cite{2021_ICLR_ViT} &
  \href{https://huggingface.co/timm/vit_large_patch16_224.augreg_in21k_ft_in1k}{vit{\_}large{\_}patch16{\_}224.augreg{\_}in21k{\_}ft{\_}in1k} \\
ViT-L/16 \cite{2021_ICLR_ViT}                & \href{https://huggingface.co/google/vit-large-patch16-224}{vit-large-patch16-224}                        \\
Swin-L \cite{2021_ICCV_Swin} &
  \href{https://huggingface.co/microsoft/swin-large-patch4-window7-224}{swin{\_}large{\_}patch4{\_}window7{\_}224.ms{\_}in22k{\_}ft{\_}in1k} \\
CSPResNet50 \cite{2020_CVPRW_Wang_CPSNet}    & \href{https://huggingface.co/timm/cspresnet50.ra_in1k}{cspresnet50}                                      \\
CSPDarkNet53 \cite{2020_arXiv_YOLOv4}        & \href{https://huggingface.co/timm/cspdarknet53.ra_in1k}{cspdarknet53.ra{\_}in1k}                         \\
NASNet-L \cite{2018_CVPR_NASNet}             & \href{https://huggingface.co/timm/nasnetalarge.tf_in1k}{nasnetalarge.tf{\_}in1k}                         \\
DarkNet53 \cite{2018_arXiv_Redmon}           & \href{https://huggingface.co/timm/darknet53.c2ns_in1k}{darknet53.c2ns{\_}in1k}                           \\
ResNet101 \cite{2016_CVPR_ResNet}            & \href{https://huggingface.co/timm/resnet101.tv_in1k}{resnet101.tv{\_}in1k}                               \\
ResNet50 \cite{2016_CVPR_ResNet}             & \href{https://huggingface.co/timm/resnet50.tv_in1k}{resnet50.tv{\_}in1k}                                 \\
VGG16 \cite{2015_ICLR_VGG}                   & \href{https://huggingface.co/timm/vgg16.tv_in1k}{vgg16.tv{\_}in1k}                                       \\
InternImage-G \cite{2023_CVPR_Wang}          & \href{https://huggingface.co/OpenGVLab/internimage_g_22kto1k_512}{internimage{\_}g{\_}22kto1k{\_}512}    \\ \bottomrule
\end{tabular}%
}
\end{table}

\begin{table}[]
\centering
\caption{Object detectors used for experiments. Model name is how the model is referred to in the text and figures. In addition, we provide the original model id and URL.}
\label{tab:obj_det}
\resizebox{0.6\columnwidth}{!}{%
\begin{tabular}{@{}ll@{}}
\toprule
Model name                                       & Model id, URL                                                                                               \\ \midrule
YOLOv11-XL                                       & \href{https://docs.ultralytics.com/models/yolov11}{yolo11x}                                                 \\
GroundingDINO B \cite{2024_ECCV_GroundingDINO}   & \href{https://huggingface.co/IDEA-Research/grounding-dino-base}{grounding-dino-base}                        \\
RT-DETR R101{\_}O365 \cite{2024_CVPR_Zhao}          & \href{https://huggingface.co/PekingU/rtdetr_r101vd_coco_o365}{rtdetr{\_}r101vd{\_}coco{\_}o365}                      \\
RT-DETR-XL \cite{2024_CVPR_Zhao}                 & \href{https://docs.ultralytics.com/models/rtdetr/}{rtdetr-x}                                                \\
RT-DETRv2 R101 \cite{2024_CVPR_Zhao}             & \href{https://huggingface.co/PekingU/rtdetr_v2_r101vd}{rtdetr{\_}v2{\_}r101vd}                                    \\
DINO Swin-L \cite{2023_ICLR_DINO}                & \href{https://github.com/IDEA-Research/DINO}{dino-swin-l}                                                   \\
DINO R50 \cite{2023_ICLR_DINO}                   & \href{https://github.com/IDEA-Research/DINO}{dino-r50}                                                      \\
OWL-ViT \cite{2022_ECCV_OWL-VIT}                 & \href{https://huggingface.co/google/owlvit-large-patch14}{owlvit-large-patch14}                             \\
MR-CNN VitDet ViT-L \cite{2022_ECCV_Li_VitDet}   & \href{https://github.com/facebookresearch/detectron2/tree/main/projects/ViTDet}{mask{\_}rcnn{\_}vitdet{\_}l}         \\
CMR-CNN VitDet Swin-L \cite{2022_ECCV_Li_VitDet} & \href{https://github.com/facebookresearch/detectron2/tree/main/projects/ViTDet}{cascade{\_}mask{\_}rcnn{\_}swin{\_}l}   \\
CMR-CNN VitDet ViT-L \cite{2022_ECCV_Li_VitDet}  & \href{https://github.com/facebookresearch/detectron2/tree/main/projects/ViTDet}{cascade{\_}mask{\_}rcnn{\_}vitdet{\_}l} \\
DETA Swin-L \cite{2022_arXiv_Zhang}              & \href{https://huggingface.co/jozhang97/deta-swin-large}{deta-swin-large}                                    \\
DETA R50 \cite{2022_arXiv_Zhang}                 & \href{https://huggingface.co/jozhang97/deta-resnet-50}{deta-resnet-50}                                      \\
DETR R50 \cite{2020_ECCV_Carion}                 & \href{https://huggingface.co/facebook/detr-resnet-50}{detr-resnet-50}                                       \\
RetinaNet R50 \cite{2016_ECCV_Liu} &
  \href{https://pytorch.org/vision/main/models/generated/torchvision.models.detection.retinanet_resnet50_fpn_v2.html}{RetinaNet{\_}ResNet50{\_}FPN} \\
Faster R-CNN R50 \cite{2015_ICCV_Girshik} &
  \href{https://docs.pytorch.org/vision/main/models/generated/torchvision.models.detection.fasterrcnn_resnet50_fpn.html}{FasterRCNN{\_}ResNet50{\_}FPN} \\
YOLOv12-XL                                       & \href{https://docs.ultralytics.com/models/yolo12/}{yolo12x}                                                 \\ \bottomrule
\end{tabular}%
}
\end{table}

\begin{table}[]
\centering
\caption{VLMs used for experiments. Model name is how the model is referred to in the text and figures. In addition, we provide the original model id and URL.}
\label{tab:vlm}
\resizebox{0.6\columnwidth}{!}{%
\begin{tabular}{@{}ll@{}}
Model name                                     & Model id, URL                                                                           \\
Qwen2-VL-7B \cite{2024_arXiv_Qwen2VL}          & \href{https://huggingface.co/Qwen/Qwen2-VL-7B-Instruct}{Qwen2-VL-7B-Instruct}       \\
Qwen2-VL-2B \cite{2024_arXiv_Qwen2VL}          & \href{https://huggingface.co/Qwen/Qwen2-VL-2B-Instruct}{Qwen2-VL-2B-Instruct}       \\
LLaVa1.6-7B \cite{2024_CVPR_LLaVa}             & \href{https://huggingface.co/liuhaotian/llava-v1.6-vicuna-7b}{llava-v1.6-vicuna-7b} \\
VILA1.5-3B \cite{2024_CVPR_VILA}               & \href{https://huggingface.co/Efficient-Large-Model/VILA1.5-3b}{VILA1.5-3b}          \\
VILA1.5-7B \cite{2024_CVPR_VILA}               & \href{https://huggingface.co/Efficient-Large-Model/VILA1.5-7b}{VILA1.5-7b}          \\
InternVL2-2B \cite{2024_arXiv_InternVL2}       & \href{https://huggingface.co/OpenGVLab/InternVL2-2B}{InternVL2-2B}                  \\
InternVL2-4B \cite{2024_arXiv_InternVL2}       & \href{https://huggingface.co/OpenGVLab/InternVL2-4B}{InternVL2-4B}                  \\
InternvL2-8B \cite{2024_arXiv_InternVL2}       & \href{https://huggingface.co/OpenGVLab/InternVL2-8B}{InternVL2-8B}                  \\
PaliGemma-3B@448px \cite{2024_arXiv_PaliGemma} & \href{https://huggingface.co/google/paligemma-3b-mix-448}{paligemma-3b-mix-448}     \\
PaliGemma-3B@224px \cite{2024_arXiv_PaliGemma} & \href{https://huggingface.co/google/paligemma-3b-mix-224}{paligemma-3b-mix-224}     \\
DeepSeek-VL-1.3B \cite{2024_arXiv_DeepSeekVL} & \href{https://huggingface.co/deepseek-ai/deepseek-vl-1.3b-chat}{DeepSeek-VL-1.3B-chat} \\
DeepSeek-VL-7B \cite{2024_arXiv_DeepSeekVL}    & \href{https://huggingface.co/deepseek-ai/deepseek-vl-7b-chat}{DeepSeek-VL-7B-chat} 
\end{tabular}%
}
\end{table}

\section{Evaluating VLMs}
\label{suppl:evaluation}

Although the expected answer for all of our questions is one word or one digit, computing accuracy for VLMs is challenging because many of them do not follow the prompt correctly. Simple matching of the keywords over-inflates accuracy because it registers a hit when the prompt is returned as part of the answer or when models consider each option in order for chain of reasoning. We manually cleaned the data to bring answers of all models to the common form. Specifically, we converted all all spelled-out numeric answers to ints (e.g. ``three'' $\rightarrow$ 3) and used regular expressions to remove common expressions (e.g. ``The objects in this image are''). These simple operations cleaned up nearly 80\% of the data. We examined the remaining 20\% and extracted the answers by hand.

We then estimated factual errors as mismatch between the answers and the ground truth. It should be noted that this approach likely inflates the accuracy. In many cases, the models output extra information, not mentioned in the plot. For example, for the open-ended categorization question some models also mention the count of objects (e.g. they answer ``three cups'' instead of ``cup''). For these answers, we only checked that the question in the query was answered (e.g. objects were correctly identified as cups) but not whether the additional information was correct (e.g. there could be fewer or more cups in the image).

To estimate faithfulness, we made the following checks: 1) the length of the answer (in characters) is equal or shorter than average length ground truth for that question (e.g. for counting questions, we expect at most 4 characters because all valid answers are numbers between 2 and 5); 2) the answer option exists (e.g. E) 10 would be invalid for the multi-choice counting question because the only options are A), B), C), and D)). 

Hard accuracy score for each image-question pair is assigned 1 if answer is both faithful and factual and 0 otherwise.

\section{Additional experiment results}

\begin{figure}
\centering
\begin{subfigure}[b]{0.43\textwidth}
\includegraphics[width=\textwidth]{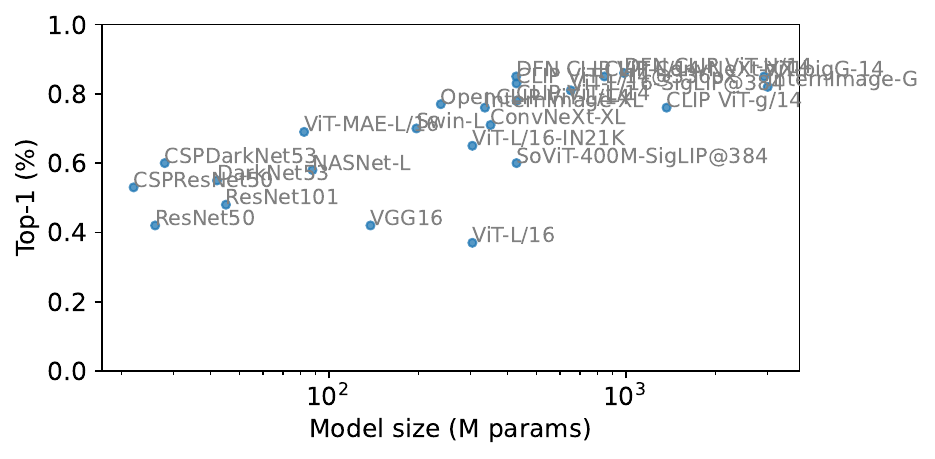}
\caption{Top-1 acc vs. model size (M params).}
\label{fig:image_classification_model_size_vs_top1_acc}
\end{subfigure}
\hspace{1em}
\begin{subfigure}[b]{0.47\textwidth}
\includegraphics[width=\textwidth]{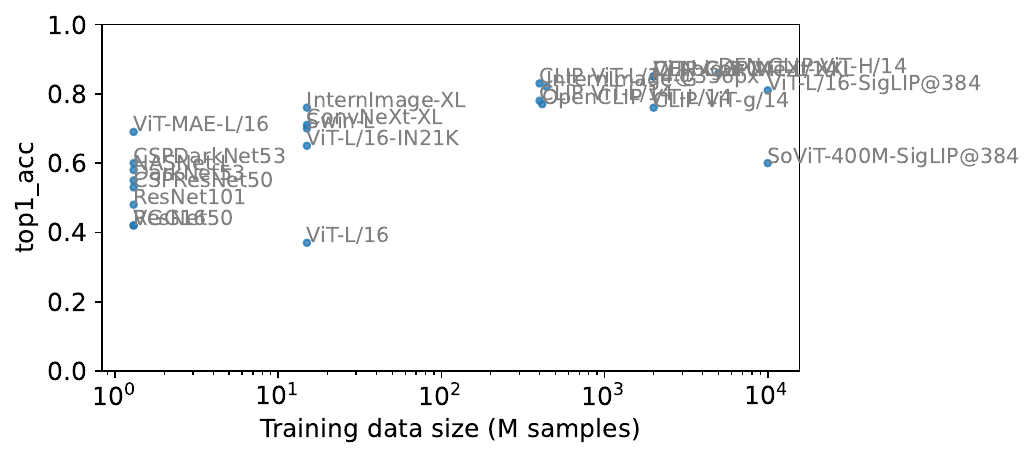}
\caption{Top-1 acc vs. data size (M samples).}
\label{fig:image_classification_data_size_vs_top1_acc}
\end{subfigure}

\begin{subfigure}[b]{0.6\textwidth}
\includegraphics[width=\textwidth]{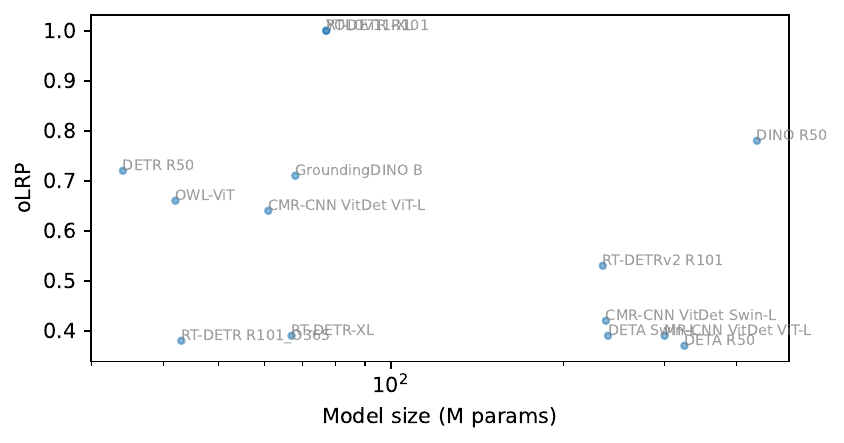}
\caption{oLRP vs. model size (M params).}
\label{fig:object_detection_model_size_vs_oLRP}
\end{subfigure}

\caption{Performance of image classification and object detection models on SNAP relative to their size and training data.}
\label{fig:performance_vs_size}
\end{figure}

\begin{figure}
\centering
\includegraphics[width=0.7\textwidth]{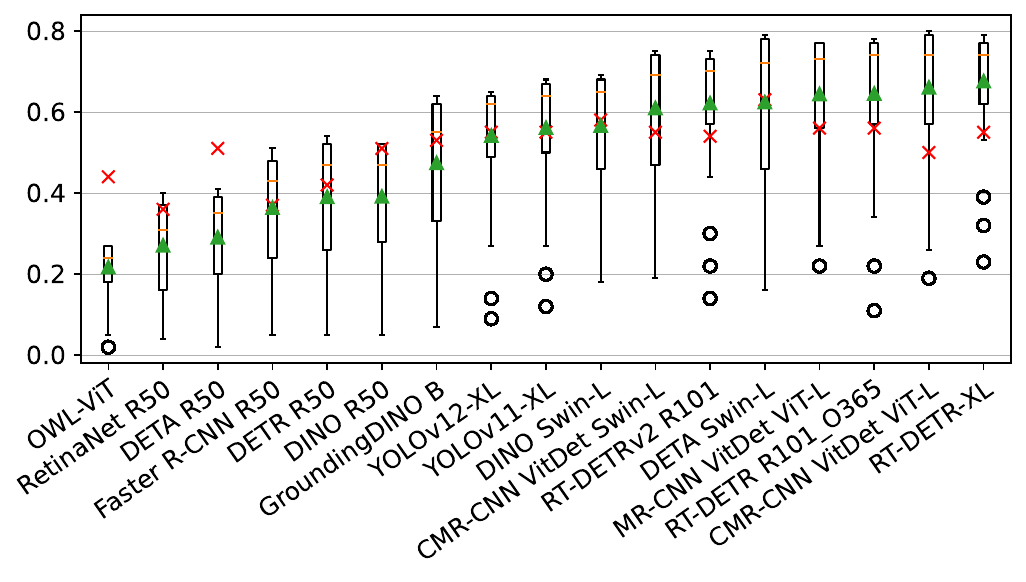}
\caption{Box plots show range and mean AP@[0.5:0.95] values for all object detection models evaluated on SNAP. Red crosses mark AP@[0.5:0.95] of the models on COCO.}
\label{fig:object_detection_EV_offset_AP_box}
\end{figure}

\begin{figure}
\centering
\begin{subfigure}[b]{0.32\textwidth}
\centering
\includegraphics[width=\textwidth]{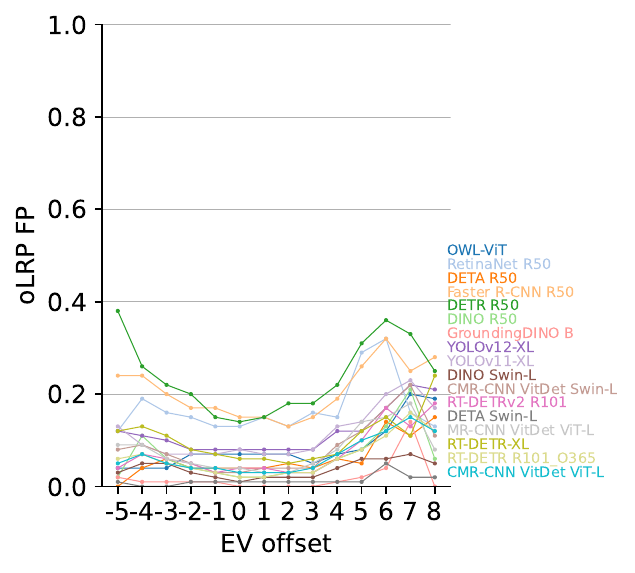}
\caption{False positive errors.}
\label{fig:obj_det_EV_offset_oLRP_FP_line}
\end{subfigure}
\begin{subfigure}[b]{0.32\textwidth}
\centering
\includegraphics[width=\textwidth]{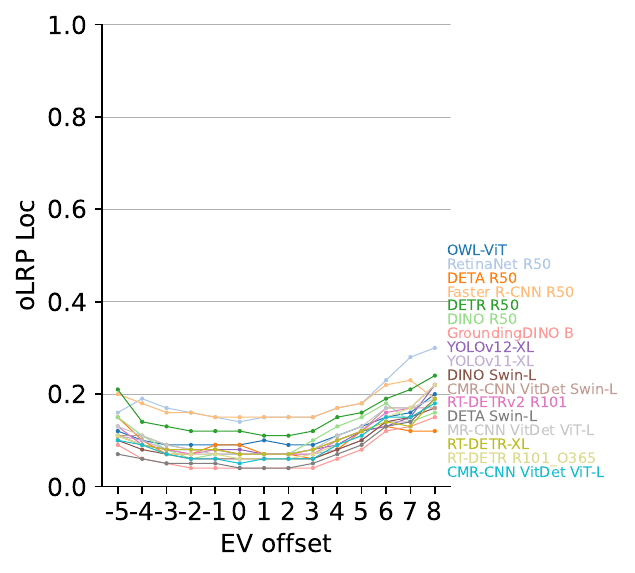}
\caption{Localization errors.}
\label{fig:obj_det_EV_offset_oLRP_Loc_line}
\end{subfigure}
\begin{subfigure}[b]{0.32\textwidth}
\centering
\includegraphics[width=\textwidth]{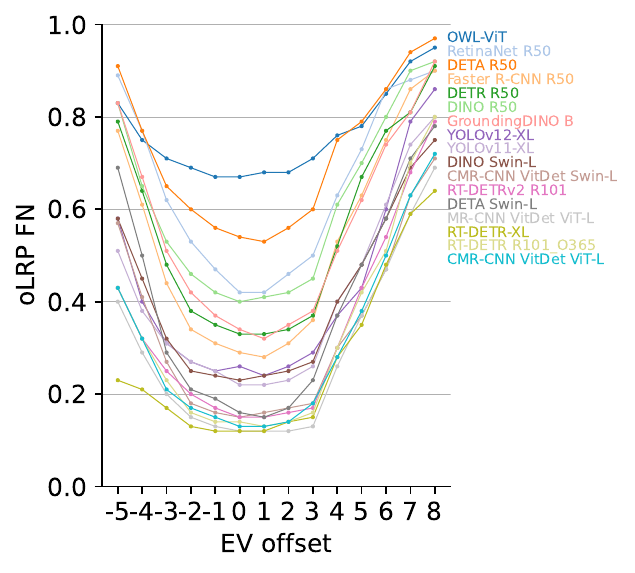}
\caption{False negative errors.}
\label{fig:obj_det_EV_offset_oLRP_FN_line}
\end{subfigure}
\caption{Classification and localization errors for all exposure bins in SNAP. Each point represents a mean of the respective error for the given EV offset.}
\label{fig:obj_det_LRP}
\end{figure}

\begin{figure}
\centering
\begin{subfigure}[b]{0.45\textwidth}
\centering
\includegraphics[width=\textwidth]{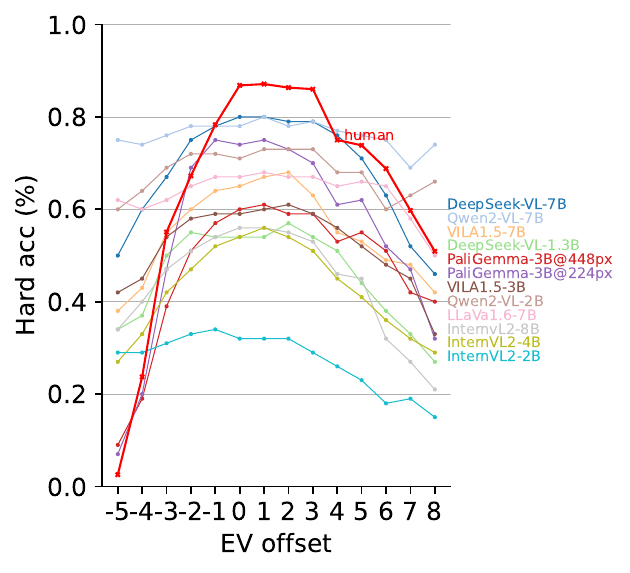}
\caption{Q1: Categorization, open-ended}
\label{fig:vqa_EV_offset_A3_acc1_line}
\end{subfigure}
\hspace{-1em}
\begin{subfigure}[b]{0.45\textwidth}
\centering
\includegraphics[width=\textwidth]{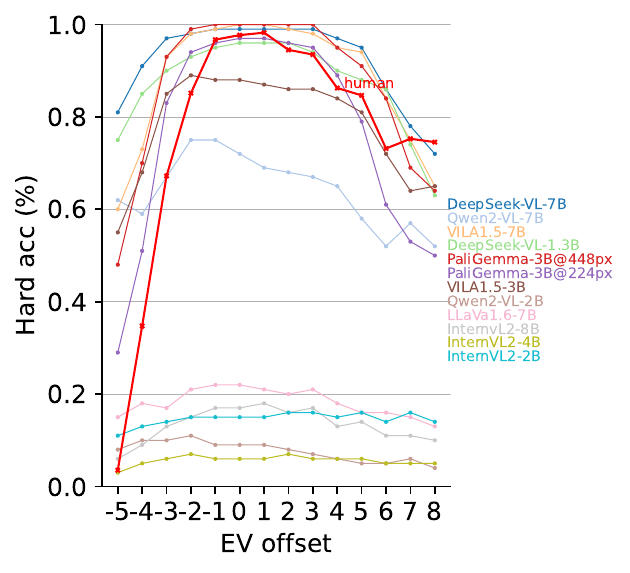}
\caption{Q3: Categorization, multiple-choice}
\label{fig:vqa_EV_offset_A5_acc1_line}
\end{subfigure}
\hspace{-1em}
\begin{subfigure}[b]{0.45\textwidth}
\centering
\includegraphics[width=\textwidth]{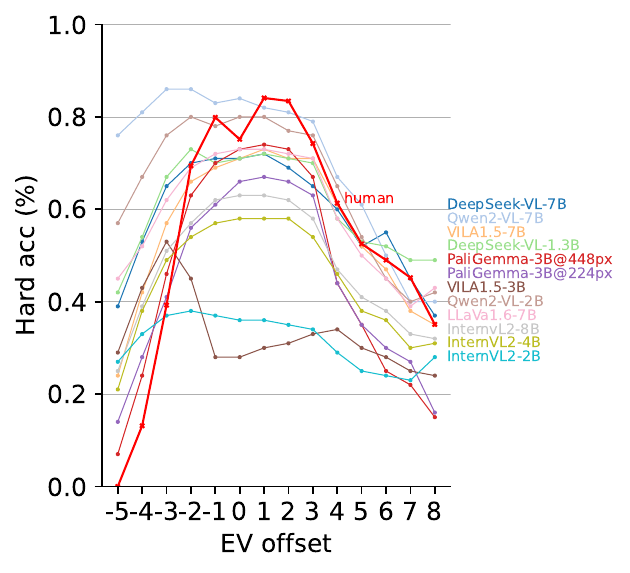}
\caption{Q2: Subitization, open-ended}
\label{fig:vqa_EV_offset_A4_acc1_line}
\end{subfigure}
\hspace{-1em}
\begin{subfigure}[b]{0.45\textwidth}
\centering
\includegraphics[width=\textwidth]{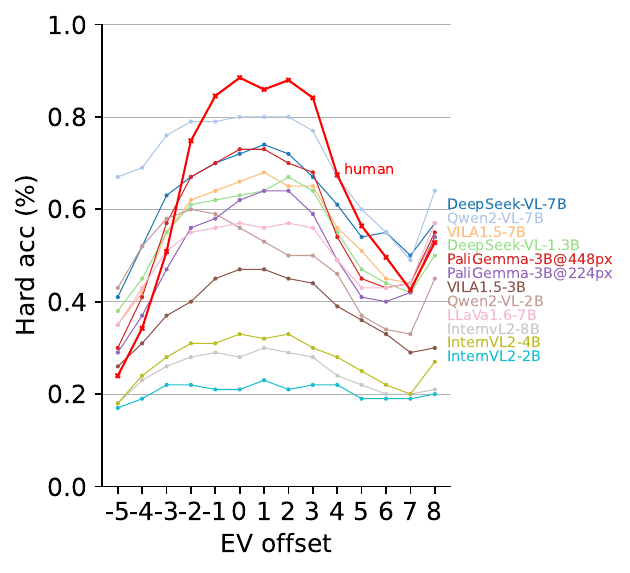}
\caption{Q4: Categorization, multiple-choice}
\label{fig:vqa_EV_offset_A6_acc1_line}
\end{subfigure}
\caption{Mean hard accuracy of VLMs across EV offsets for each VQA question.}
\label{fig:vqa_all_acc1}
\end{figure}

\begin{figure}
\centering
\begin{subfigure}[b]{0.45\textwidth}
\centering
\includegraphics[width=\textwidth]{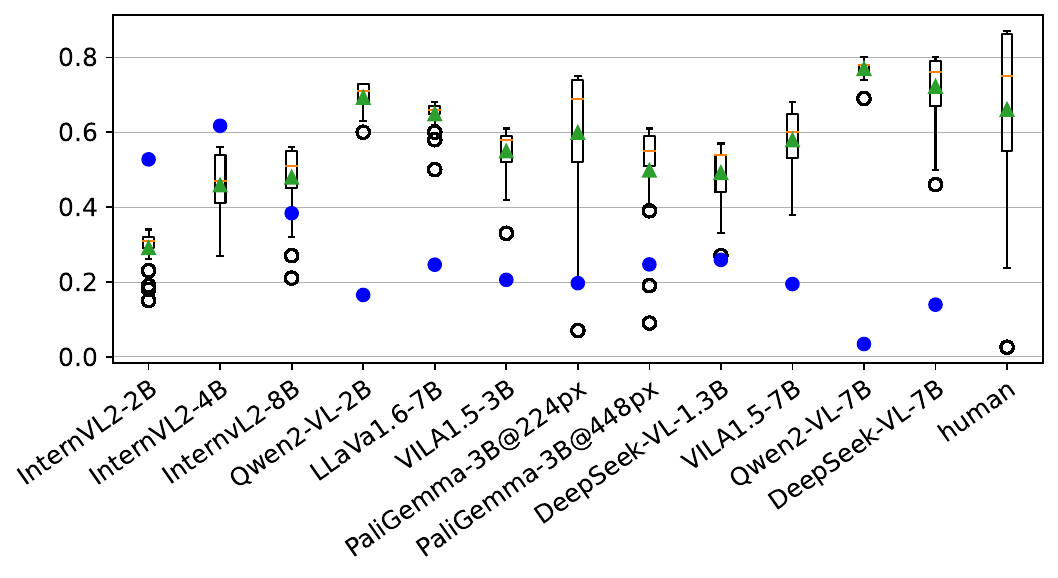}
\caption{Q1: Categorization, open-ended}
\label{fig:vqa_EV_offset_A3_acc1_box}
\end{subfigure}
\hspace{-1em}
\begin{subfigure}[b]{0.45\textwidth}
\centering
\includegraphics[width=\textwidth]{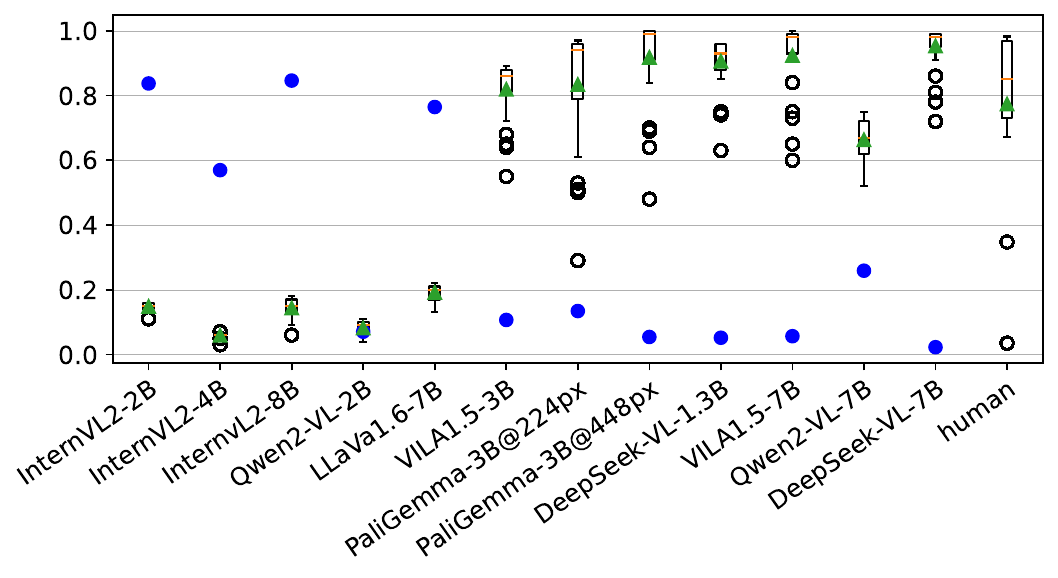}
\caption{Q3: Categorization, multiple-choice}
\label{fig:vqa_EV_offset_A5_acc1_box}
\end{subfigure}
\hspace{-1em}
\begin{subfigure}[b]{0.45\textwidth}
\centering
\includegraphics[width=\textwidth]{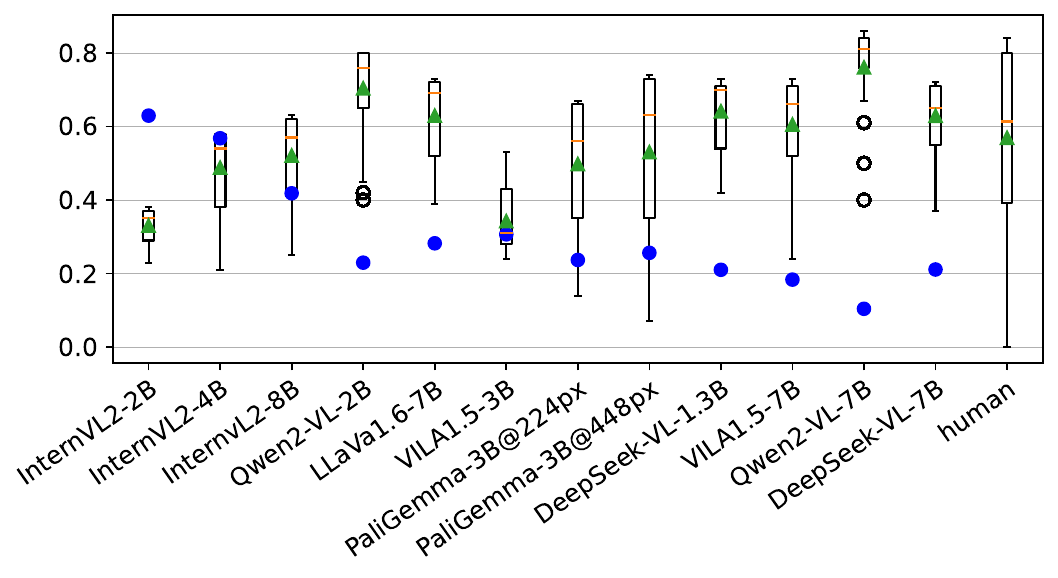}
\caption{Q2: Subitization, open-ended}
\label{fig:vqa_EV_offset_A4_acc1_box}
\end{subfigure}
\hspace{-1em}
\begin{subfigure}[b]{0.45\textwidth}
\centering
\includegraphics[width=\textwidth]{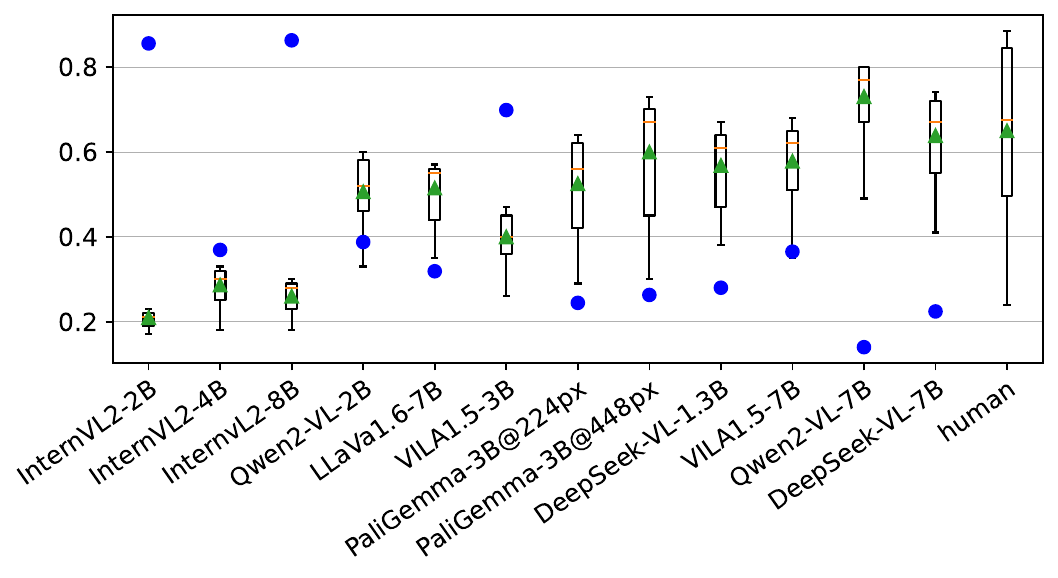}
\caption{Q4: Categorization, multiple-choice}
\label{fig:vqa_EV_offset_A6_acc1_box}
\end{subfigure}
\caption{Box plots showing the range of hard accuracy of VLMs on each question. Parameter sensitivity (PS) is marked with blue circles.}
\label{fig:vqa_all_box_acc1}
\end{figure}

In this section we provide additional experimental results. \cref{fig:performance_vs_size} contains plots that show positive correlation between performance of image classification and object detection models on SNAP and scale of the data/models.  In \cref{fig:object_detection_EV_offset_AP_box} we show that object detectors achieve better standard AP metric on SNAP than on COCO likely due to relatively simple design of SNAP, which lacks significant clutter and occlusions. Additional results for object detection task in \cref{fig:obj_det_LRP} shows a larger contribution of false negatives relative to false positives and localization errors. Because FN errors mimic patterns observed in image classification task, we speculate that the likely cause of this is the influence of image classification backbones. Lastly, \cref{fig:vqa_all_acc1} shows hard accuracy of VLMs across exposures w.r.t. to human baseline and \cref{fig:vqa_all_box_acc1} shows parameter sensitivity  for each question in VQA task. 

\end{document}